%% file: main.tex
\documentclass[10pt,twocolumn,letterpaper]{article}

\usepackage[pagenumbers]{cvpr} 

\input{preamble}

\definecolor{cvprblue}{rgb}{0.21,0.49,0.74}
\usepackage[pagebackref,breaklinks,colorlinks,citecolor=cvprblue]{hyperref}

\def\paperID{9823} 
\def\confName{CVPR}
\def\confYear{2024}

\title{LiDAR-based Person Re-identification}

\author{Wenxuan Guo\textsuperscript{1}
	\qquad Zhiyu Pan\textsuperscript{1}
	\qquad Yingping Liang\textsuperscript{2}
	\qquad Ziheng Xi\textsuperscript{1}\\
	\qquad Zhicheng Zhong\textsuperscript{1}
	\qquad Jianjiang Feng\textsuperscript{1}\thanks{Corresponding author}
	\qquad Jie Zhou\textsuperscript{1}\\
	\textsuperscript{1}Tsinghua University \qquad \textsuperscript{2}Beijing Institute of Technology\\
	{\tt\small\{gwx22,pzy20,xizh21,zhongzc18\}@mails.tsinghua.edu.cn \qquad liangyingping@bit.edu.cn} \\
	{\tt\small \{jfeng,jzhou\}@tsinghua.edu.cn}}

\begin{document}
\maketitle
\input{sec/0_abstract}    
\input{sec/1_intro}

\input{sec/2_related}

\input{sec/4_method}

\input{sec/3_dataset}

\input{sec/5_experiment}
\input{sec/6_conclu}

\clearpage

{
	\small
	\bibliographystyle{ieeenat_fullname}
	\bibliography{main}
}
\input{sec/X_suppl}

\end{document}

%% file: preamble.tex
\usepackage[dvipsnames]{xcolor}

\usepackage{fontawesome}
\usepackage{utfsym}
\usepackage{multirow}
\usepackage{threeparttable}
\usepackage{svg}

%% file: sec/0_abstract.tex
\begin{abstract}
Camera-based person re-identification (ReID) 
systems have been widely applied in the field of public security. 
However, cameras often lack the perception of 3D morphological information of human and are susceptible to various limitations, 
such as inadequate illumination, complex background, and personal privacy. 
In this paper, we propose a LiDAR-based ReID framework, ReID3D, 
that utilizes pre-training strategy to retrieve features of 3D body shape and introduces 
Graph-based Complementary Enhancement Encoder for extracting comprehensive features. 
Due to the lack of LiDAR datasets, we build LReID, the first LiDAR-based person ReID dataset, 
which is collected in several outdoor scenes with variations in natural conditions. Additionally, 
we introduce LReID-sync, a simulated pedestrian dataset designed for pre-training encoders with tasks of point cloud completion 
and shape parameter learning. 
Extensive experiments on LReID show that ReID3D achieves exceptional performance with a rank-1 accuracy of 94.0, 
highlighting the significant potential of LiDAR in addressing person ReID tasks. To the best of our knowledge, we are the first to propose a solution for LiDAR-based ReID. Code is available at \href{https://github.com/GWxuan/ReID3D}{https://github.com/GWxuan/ReID3D}.
\end{abstract}

%% file: sec/1_intro.tex
\section{Introduction}
\label{sec:intro}

Person ReID has numerous practical applications, such as video surveillance, intelligent transportation and public security. Most ReID systems use cameras as sensors, aiming to recognize the same individual in images or videos caught by different cameras. 
With the development of computer vision technology, camera-based ReID has witnessed continuous advancements. However, certain challenges have not yet been effectively addressed. Firstly, cameras introduce limitations in terms  of visual ambiguity caused by poor illumination and complex backgrounds~\cite{shen2023lidargait}. Additionally, current camera-based ReID models primarily learn appearance information~\cite{yao2019deep}. Therefore, variations in human appearance considerably impact the performance of models~\cite{leng2019survey}. Furthermore, camera-based person ReID systems raise personal privacy issues~\cite{ahmed2020camera,brkic2017face} for applications in some areas.

\begin{figure}
	\centering
	\includegraphics[width=0.45\textwidth]{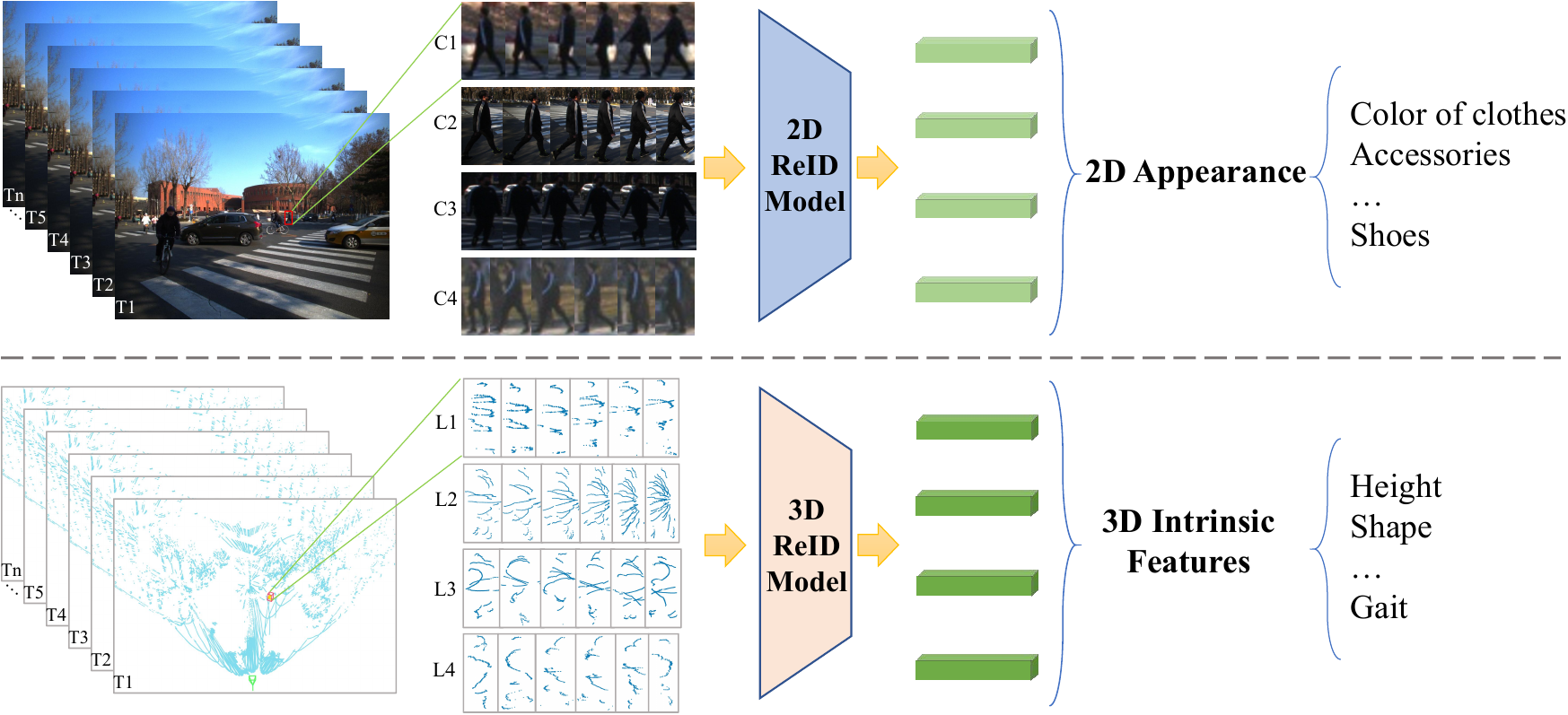}
	\caption{Overview of camera-based and LiDAR-based person ReID. Camera-based methods primarily learn the 2D appearance of human, such as color of clothes, accessories, and shoes. LiDAR-based methods utilize 3D structural information to learn intrinsic features, such as height, body shape, and gait.}
	\vspace{-.2cm}
	\label{fig:intro}
	\vspace{-.2cm}
\end{figure}

In recent years, Kinect-based ReID approaches have been proposed~\cite{munsell2012person,ren2017multi} to address these challenges by utilizing depth information. However, Kinect is primarily designed for indoor scenes and has a limited measurement range~\cite{lachat2015first}, restricting its applicability in large-scale outdoor scenarios. Recently, researchers have explored the use of radar for gait recognition, identity recognition, and person ReID~\cite{meng2020gait, cheng2021person}. While radar is cost-effective, it suffers from poor angular and distance resolution, as well as a limited effective measurement range~\cite{cheng2021person}, which poses challenges in discriminating individuals with similar body shapes.

In the past few years, LiDAR has been widely adopted in autonomous driving, driven by its improved accuracy and reduced costs~\cite{wang2019pseudo}. The successful applications of LiDAR 
motivate us to explore its potential in person ReID within complex outdoor scenes. 
LiDAR can 
offer a broader measurement range and higher resolution, enabling precise perception about individuals in large-scale outdoor settings.   
LiDAR provides precise 3D structural information  without being affected by lighting conditions or complex backgrounds, which enables the extraction of intrinsic features of individuals, such as height, body shape, and gait, irrespective of clothing color.
LiDAR has been utilized in the task of gait recognition~\cite{shen2023lidargait}, which focuses on recognizing cooperative subjects walking along a specified route in small-scale scenes. However, ReID primarily focuses on the identification of unscripted pedestrians in large-scale scenes with a small sample size for each identity, which often requires multiple acquisition devices. 
In our work, we leverage low-cost LiDARs to tackle person ReID challenges, as shown in Fig.~\ref{fig:intro}.

To the best of our knowledge, this paper presents the first study on LiDAR-based person ReID. We build LReID, the first LiDAR-based ReID dataset to facilitate research on utilizing LiDAR point clouds for person ReID. We collect the dataset in several outdoor scenes using multiple collection nodes, each including a Livox Mid-100 LiDAR and an industrial camera. LReID offers several distinctive features: (1) \textbf{Real-world scenes}. The dataset is captured in outdoor scenes where pedestrians demonstrate unscripted behavior, leading to occlusions between pedestrians, as well as the presence of dynamic objects like vehicles and bicycles that may impact person ReID. (2) \textbf{Data diversity}. LReID encompasses dynamic data and annotations of 320 pedestrian gathered in varying seasons, times of day, and lighting conditions, amounting to a total of 156,000 frames of point clouds and images, thus enabling comprehensive analysis of the impact of different factors on person ReID. (3) \textbf{Precision}. The Livox Mid-100 LiDAR has a distance accuracy of 2 cm and an angular accuracy of 0.1°, which provides high-precision 3D structural information for ReID problems. 

Additionally, we introduce a simulated dataset, named LReID-sync, including 360,000 frames of point clouds for 600 pedestrians captured by multi-view synchronous LiDARs. LReID-sync comprises annotations for point cloud completion from single view to full views, as well as Skinned Multi-Person Linear Model (SMPL) parameters~\cite{SMPL2015}.

Based on point clouds, the identification of pedestrians relies on their static anthropometric features, including height, body shape, and limb structure, as well as their dynamic gait features. Accurately extracting complete shape features of a pedestrian is beneficial for both aspects. To address this, we propose an efficient LiDAR-based framework, termed ReID3D.
ReID3D utilizes a pre-training strategy to guide the encoder in learning 3D body features based on LReID-sync. Moreover, in order to extract discriminative static and dynamic features of pedestrians, the ReID network of ReID3D comprises a Graph-based Complementary Enhancement Encoder (GCEE) and a temporal module. Extensive experiments on LReID demonstrate the following: (1) ReID3D outperforms the state-of-the-art camera-based methods, particularly under low light, highlighting the significant potential of LiDAR in addressing person ReID tasks. (2) The use of LReID-sync for pre-training significantly enhances feature encoding capability of the model. (3) Compared to commonly used point cloud encoders, our GCEE exhibits a higher proficiency in extracting comprehensive and discriminative features.

To summarize, our main contributions are as follows:
\begin{itemize}
	\item[$\bullet$] To the best of our knowledge, this is the first work on LiDAR-based person ReID, demonstrating the practicality of utilizing LiDAR for person ReID in challenging real-world outdoor scenes.
	\item[$\bullet$] We build LReID, the first LiDAR-based person ReID dataset, which is collected in several outdoor scenes with variations in natural conditions. Moreover, we introduce LReID-sync, a new simulated pedestrian dataset designed for pre-training ReID models with tasks of point cloud completion and shape parameter learning.
	\item[$\bullet$] We propose a LiDAR-based ReID framework, termed ReID3D, that utilizes pre-training strategy to guide the encoder in learning 3D body features and introduces GCEE for extracting comprehensive and discriminative features. Experimental results on dataset LReID indicate that ReID3D outperforms camera-based methods.
\end{itemize}

%% file: sec/2_related.tex
\section{Related Work}
\label{sec:related}

\begin{table*}[]
	\footnotesize
	\centering
	\caption{Comparison of publicly available 3D datasets for person ReID.}\label{tab:dataset}
	\vspace{-.15cm}
	\begin{tabular}{cccccccc}
		\hline 
		Dataset & Year & Identity & Sensor & Scene & Unscripted & Natural diversity&Occlusion\\ \hline
		RGBD-ID~\cite{rgbdreid}& 2012 & 80 & 1 Kinect &Indoor &\textcolor{red}{\usym{2717}}&\textcolor{red}{\usym{2717}}&\textcolor{red}{\usym{2717}}\\
		BIWI RGBD-ID~\cite{biwi}& 2014 & 50 & 1 Kinect & Indoor &\textcolor{red}{\usym{2717}}&\textcolor{red}{\usym{2717}}&\textcolor{red}{\usym{2717}}\\
		Kinect-REID~\cite{kinectreid}& 2015 & 71 & 1 Kinect & Indoor &\textcolor{red}{\usym{2717}}&\textcolor{red}{\usym{2717}}&\textcolor{red}{\usym{2717}}\\\hline
		\textbf{LReID}& 2023 & \textbf{320} & \textbf{4 LiDARs + 4 Cameras} & \textbf{Outdoor} &\textcolor{green}{\usym{2714}}&\textcolor{green}{\usym{2714}}&\textcolor{green}{\usym{2714}}\\
		\textbf{LReID-sync}& 2023 & \textbf{600} & \textbf{4 LiDARs} & Simulation &\textcolor{green}{\usym{2714}}&\textcolor{red}{\usym{2717}}&\textcolor{green}{\usym{2714}}\\\hline
	\end{tabular}
	\vspace{-.15cm}
\end{table*}

\paragraph{Person ReID.}
Camera-based ReID has been extensively researched in the past decades. Researchers have employed various convolutional neural networks~\cite{ye2021deep} to extract human features from images or videos. Some video-based ReID models adopt recurrent neural networks (RNNs)~\cite{mclaughlin2016recurrent,yan2016person,zhou2017see} or transformer blocks~\cite{chen2022rest,zheng2022template} to aggregate temporal features. These models primarily focus on extracting appearance features,
which serve as efficient but short-lived identifiers. By contrast, intrinsic and behavioral features exhibit minimal change over time. Given the constraints of using cameras at night, infrared sensors have been introduced for RGB-IR cross-modality person ReID~\cite{wu2017rgb,nguyen2017person}. However, the infrared sensor only captures 2D single-channel intensity information, constraining its perceptive ability.

In order to extract intrinsic features of individuals, researches on ReID utilizing Kinect~\cite{munsell2012person,ren2017multi} and radar~\cite{cheng2021person} have emerged. Kinect and radar provide 3D structural information to help the model extract physiological and behavioral features, reducing reliance on the appearance of individuals~\cite{ren2017multi,fan2020learning}. However, they are constrained by significant hardware limitations~\cite{lachat2015first,cheng2021person}. 
In comparison, LiDAR serves as a more advantageous option, offering superior practicality. Despite this, LiDAR-based person ReID has not been researched. 

\paragraph{Point Cloud Completion.}
Point cloud completion is the task to predict missing parts based on the rest of the point cloud. To accomplish this, networks need to learn the intrinsic geometric structures and semantic knowledge of the 3D object. Additionally, the learned representations can be transferred to downstream tasks. The entire process does not require human annotations and therefore falls under the category of unsupervised representation learning~\cite{xiao2023unsupervised}.

Point cloud completion has received increasing attention in the past decade~\cite{wen2020point,huang2020pf,groueix2018papier,liu2020morphing}. In terms of completing the missing parts caused by a single viewpoint, Wang \etal~\cite{wang2021unsupervised} utilize an encoder-decoder model to recover the occluded points. However, there has been limited research focusing on point cloud completion for pedestrians.

\paragraph{Person ReID Dataset.}
Depending on the used representations, person ReID datasets can be classified into 2D and 3D datasets. Due to the early emergence and widespread use of cameras, 2D datasets are mainly composed of camera-based datasets~\cite{87,88,14,19,zheng2016mars,wang2014person}, which have advanced research in person ReID. In addition, some 2D datasets utilize infrared cameras as sensors~\cite{nguyen2017person,wu2017rgb}. Kinect-based datasets~\cite{biwi,kinectreid,rgbdreid} serve as typical examples of 3D datasets. However, the current Kinect datasets have limited scales as they are often collected in small indoor scenes with only one cooperative subject. Additionally, Cheng and Liu~\cite{cheng2021person} collected a radar dataset containing 40 identities, but it is not publicly available and also involves only one cooperative subject in each frame. To address these limitations, we introduce a novel LiDAR-based 3D dataset. The comparison of publicly available 3D datasets is shown in Tab.~\ref{tab:dataset}.

%% file: sec/4_method.tex
\section{Method}
\label{sec:method}
\begin{figure}
	\centering
	\includegraphics[width=0.9\linewidth]{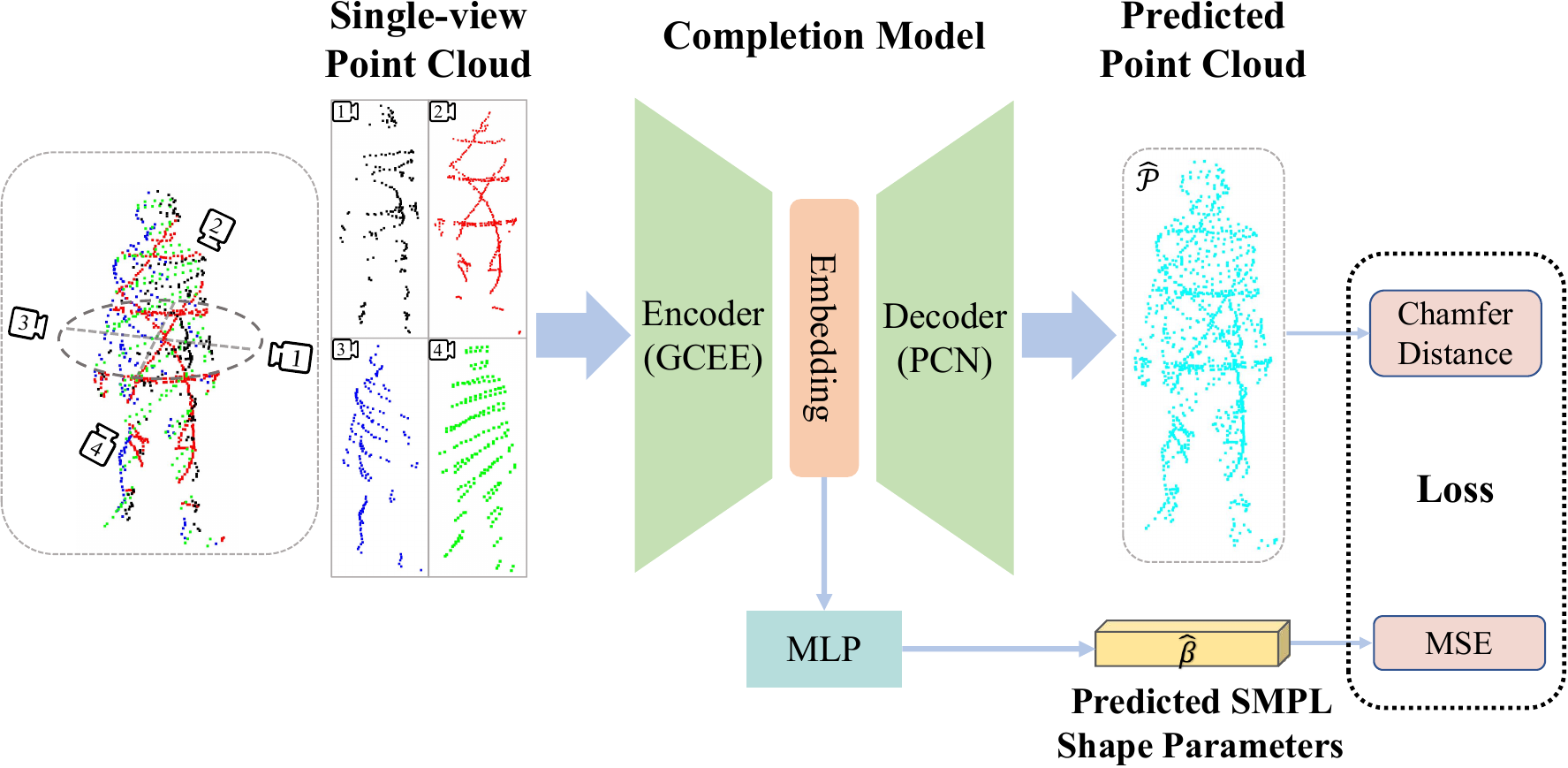}
	\vspace{-.1cm}
	\caption{The pre-training approach of ReID3D. Simulated single-view point cloud is taken as input, and the encoder is pre-trained for tasks of point cloud completion and SMPL parameter learning.}
	\label{fig:method1}
	\vspace{-.2cm}
\end{figure}

\begin{figure*}
	\centering
	\includegraphics[width=0.8\linewidth]{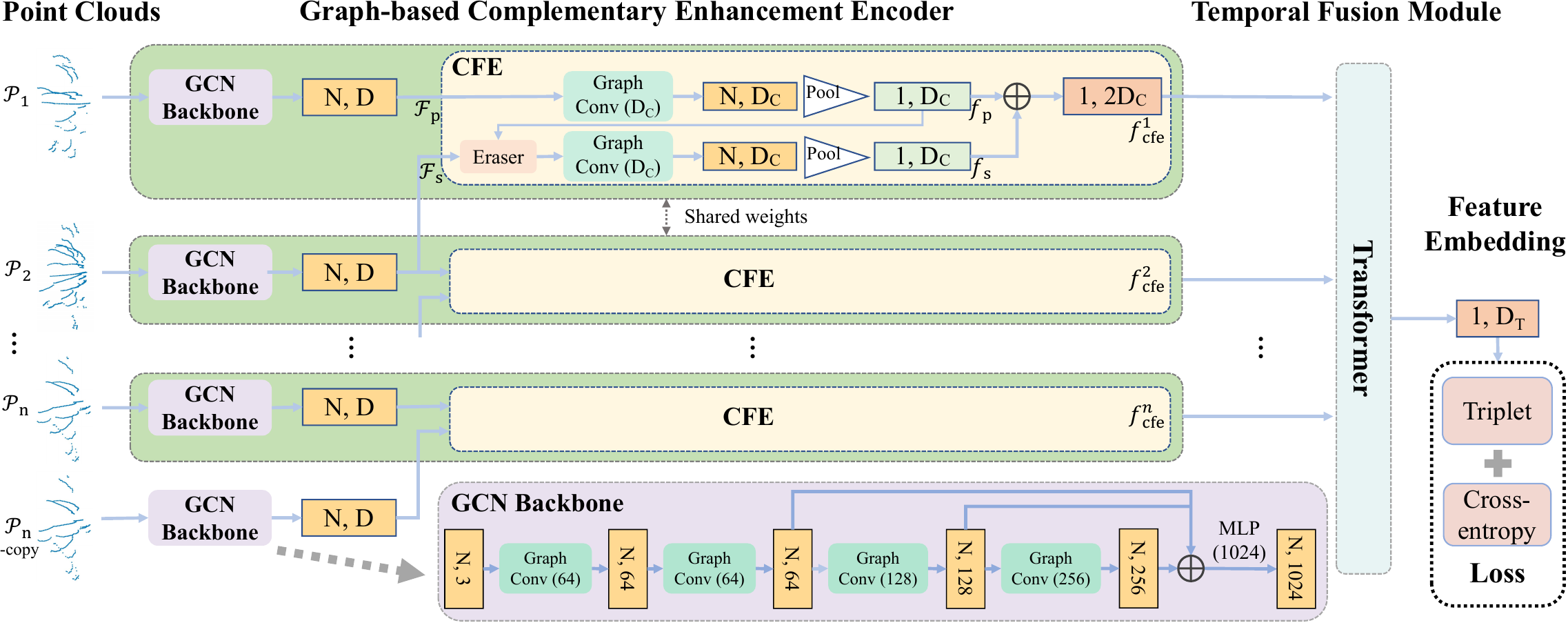}
	\caption{The ReID network of ReID3D. The Graph-based Complementary Enhancement Encoder (GCEE) extracts frame-level features from the pedestrian point clouds, and the transformer module aggregates the sequential features. GCEE consists of a GCN backbone and a Complementary Feature Extractor (CFE).}
	\label{fig:method2}
	\vspace{-.2cm}
\end{figure*}

ReID3D adopts multi-task pre-training to guide the encoder in learning 3D body features based on LReID-sync, as shown in Fig.~\ref{fig:method1}. The ReID network of ReID3D comprises a Graph-based Complementary Enhancement Encoder (GCEE), which consists of a GCN backbone and a Complementary Feature Extractor (CFE), along with a temporal module, as shown in Fig.~\ref{fig:method2}. The pre-trained GCEE is used to initialize the ReID network.

\subsection{Pre-training with Multiple Tasks}
\label{sec:pre}
Based on our observations, the crucial factors that are likely to impact the performance of ReID models are (1) the variations in information resulted from different viewpoints under cross-view settings, and (2) the incomplete information acquired from single view.
Besides, the collection and annotation of real-world data involve significant costs, while simulated data is low-cost, and comes with rich and accurate annotations. Hence, we leverage simulated data to pre-train the encoder for tasks of point cloud completion and SMPL parameter learning. The overall idea of our proposed pre-training approach is shown in Fig.~\ref{fig:method1}, which enables the encoder to effectively extract anthropometric characteristics and mitigate the influence of viewpoint disparities.

\paragraph{Multi-task Network.}
Throughout we define point cloud $\mathcal{P}$ as a set of points in 3D Euclidean space, $\mathcal{P} = \{p_{i}|i=1, 2..., N\}$, where each
point $p_{i}$ can be represented by its coordinates $(x_{i}, y_{i}, z_{i})$. The network is fed with the single-view point cloud $\tilde{\mathcal{P}}$, then embed it into a latent vector $z$ using encoder $\Phi(\cdot)$. The first branch of the network focuses on the task of point cloud completion, where the decoder $\Psi(\cdot)$ complete the point cloud based on $z$. The process can be formulated as:
\begin{equation}
	\hat{\mathcal{P}} = \Psi(\Phi(\tilde{\mathcal{P}})),
\end{equation}
where $\hat{\mathcal{P}}$ is the predicted complete point cloud.

The second branch aims to learn the characteristics of human body, \ie SMPL shape parameters $\beta$. The latent vector $z$ is expected to encompass shape information. Therefore, a straightforward MLP network is utilized to convert $z$ into predicted SMPL shape parameters $\hat{\beta}$. The process can be formulated as:
\begin{equation}
	\hat{\beta} = \text{MLP}(\Phi(\tilde{\mathcal{P}})).
\end{equation}

\paragraph{Training.}
We adapt the folding-based decoder PCN~\cite{yuan2018pcn} to complete the single-view point cloud in two steps, outputting a coarse shape $\hat{\mathcal{P}}_{\text{coarse}}$ and a detailed shape $\hat{\mathcal{P}}_{\text{detail}}$.
The pre-training model is trained with a combined loss corresponding to the two branches. In the branch of point clout completion, we use Chamfer Distance (CD) as the difference measure between prediction $\hat{\mathcal{P}}$ and ground-truth $\mathcal{Y}$:
\begin{equation}
	\text{CD}(\hat{\mathcal{P}}, \mathcal{Y}) = \frac{1}{|\hat{\mathcal{P}}|}\sum_{\hat{p} \in \hat{\mathcal{P}}} \min_{p \in \mathcal{Y}} \|\hat{p} - p\| + \frac{1}{|\mathcal{Y}|}\sum_{p \in \mathcal{Y}} \min_{\hat{p} \in \hat{\mathcal{P}}} \|p - \hat{p}\|.
\end{equation}
The loss of the completion branch is a sum of the
Chamfer distances on the coarse and detailed shapes weighted by a hyperparameter $\delta$:
\begin{equation}
	\mathcal{L}_{\text{com}} = \text{CD}(\hat{\mathcal{P}}_{\text{coarse}},\mathcal{Y}) + \delta \text{CD}(\hat{\mathcal{P}}_{\text{detail}},\mathcal{Y})\label{eq:a}.
\end{equation}
The branch of shape parameter learning utilizes mean squared error (MSE) as the loss function:
\begin{equation}
	\mathcal{L}_{\text{shape}} = \frac{1}{n} \sum_{i=1}^{n} (\beta_i - \hat{\beta_i})^2,
\end{equation}
where $n$ is the dimension of $\beta$, \ie $n=10$.

The pre-training loss is a combination of the two branches with a weighted hyperparameter $\eta$:
\begin{equation}
	\mathcal{L} = \mathcal{L}_{\text{com}} + \eta \mathcal{L}_{\text{shape}}.
\end{equation}

\subsection{ReID Network}
To extract spatio-temporal features from the sequence of point clouds, the ReID Network of ReID3D comprises a GCEE, which consists of a GCN backbone and CFE, along with a temporal module, as illustrated in Fig.~\ref{fig:method2}.

\paragraph{GCN Backbone.}
To extract local and global features from point clouds effectively, we employ graph convolutional structure as the backbone. We construct the directed graph $\mathcal{G}(\mathcal{V}, \mathcal{E})$ by using the k-nearest neighbors (KNN) of each point including self-loop, where $\mathcal{V}$ represents the points and $\mathcal{E}$ represents the set of edges. For the original point cloud, we construct the graph based on the nearest neighbors in its coordinate space. However, we update the graph at each layer of the network in feature space based on the feature similarity among points, rather than fixed spatial positions.

Denote that $p_i$ is the central point of once graph convolution operation, and $\mathcal{N}(i) = \{j | (i, j) \in \mathcal{E}\}$ is the set of points
in its neighborhood with features $\mathcal{F}(i) = \{f_{j} | j\in \mathcal{N}(i)\}$. To integrate the global shape structure and local neighborhood feature differences~\cite{wang2019dynamic}, we define $f_{ij}= [f_i, f_j - f_i]$ as the input feature, where $[\cdot, \cdot]$ is the concatenation operation. The convolution process can be formulated as:
\begin{equation}
	f_i^{'} = \max_{j\in \mathcal{N}(i)} \sigma (f_{ij} \times k),
\end{equation}
where $f_i^{'}$ is the output feature of the central point $p_i$, $k$ is the convolution kernel, $\max$ is a channel-wise max-pooling function, $\times$ represents matrix multiplication and $\sigma(\cdot)$ is the activation function LeakyReLU.

\begin{figure}
	\centering
	\includegraphics[width=0.8\linewidth]{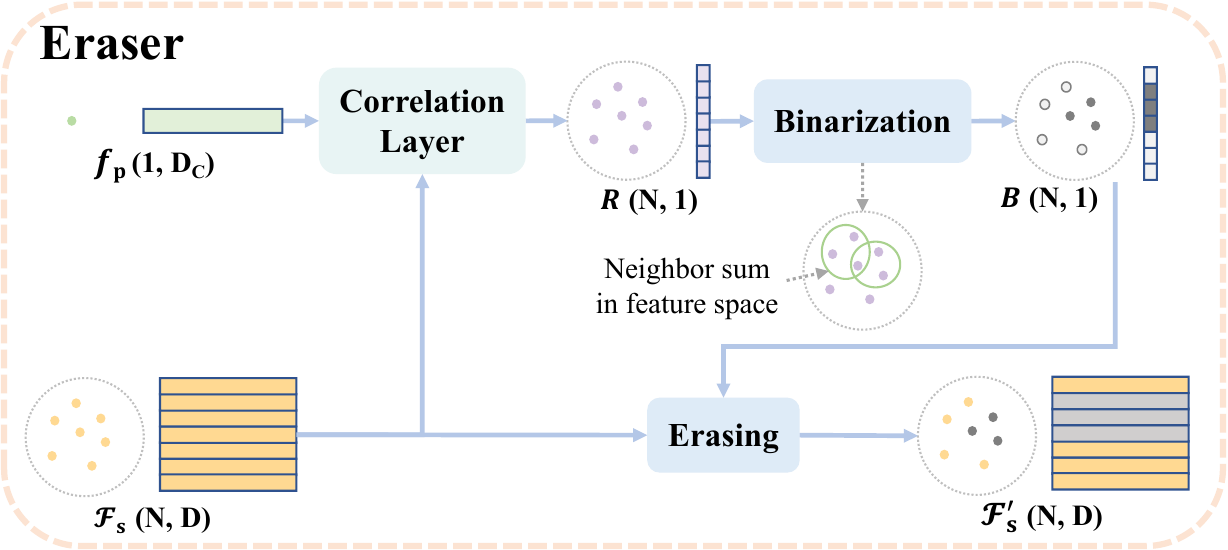}
	\caption{The architecture of the Eraser module in CFE.}
	\label{fig:method3}
\end{figure}

\paragraph{Complementary Feature Extractor.} 
One viable method for extracting frame-level pedestrian features is to apply pooling directly to the features output by the GCN backbone. However, applying the same operation to each frame results in highly redundant extraction, which only emphasizes a part of the most salient features. To this end, inspired by \cite{hou2020temporal}, we design CFE to extract complementary features and improving representation learning of the encoder.

For each frame in a sequence, we consider it as the primary frame. The CFE further extracts its features based on the output of the GCN backbone. Then, the next frame is taken as the supplementary frame, from which the Eraser module erases the previously discovered salient features. We duplicate the last frame $\mathcal{P}_n$ as its own supplementary frame. 
In particular, for a sequence of point clouds  $\{\mathcal{P}_i\}_{i=1}^n (\mathcal{P}_i \in \mathbb{R}^{N\times 3})$, the feature sequence $\{\mathcal{F}_i\}_{i=1}^n(\mathcal{F}_i \in \mathbb{R}^{N\times D})$ is obtained by the GCN backbone, where $N$ is the number of points and $D$ is the dimension of features. For the primary frame, CFE utilizes a graph convolution layer and a global max-pooling layer to extract the most salient features $f_{\text{p}} \in \mathbb{R}^{1\times D_\text{C}}$. Then, the Eraser module reconstruct the features of the supplementary frame $\mathcal{F}_\text{s}$ guided by the discovered features $f_{\text{p}}$. The reconstructed features $\mathcal{F}_\text{s}^{'}$ is fed into similar graph convolution layer and global max-pooling layer to extract auxiliary features $f_{\text{s}}$. $f_{\text{p}}$ and $f_{\text{s}}$ are concatenated to obtain the complementary features $f_{\text{cfe}} \in \mathbb{R}^{1\times 2D_\text{C}}$ of the primary frame.

The architecture of the Erasure module is shown in Fig.~\ref{fig:method3}. Firstly, the correlation layer is used to obtain the correlation vector $R \in \mathbb{R}^{N\times 1}$ between $\mathcal{F}_\text{s}$ and $f_{\text{p}}$, through computing the semantic relevance between $f_{\text{p}}$ and all the local descriptors of $\mathcal{F}_\text{s}$, formulated as:
\begin{equation}
	R = \mathcal{F}_\text{s} \times (f_{\text{p}} \times \omega)^T,
\end{equation}
where $\omega \in \mathbb{R}^{D_\text{C}\times D}$ is a learnable variable projecting $f_\text{p}$ to the feature space of $\mathcal{F}_\text{s}$. 
Then, the Binarization module generates the binary mask based on $R$ to identify the points to be erased. Specifically, for each point in $\mathcal{F}_\text{s}$, calculate the sum of the correlation values of the point and its $K_\text{B}$ nearest neighbors in the feature space. The calculated sum represents the correlation of the region centered on the point. We select the region with the highest correlation value to be erased, through setting the corresponding value in the binary mask $B$ to 0 and others to 1. Finally, $\mathcal{F}_\text{s}$ is erased according to $B$ to obtain the reconstructed features $\mathcal{F}_\text{s}^{'}$. 

\paragraph{Temporal Fusion Module.}
The sequence of pedestrian point clouds includes distinctive dynamic features, such as gait frequency and amplitude of limb swings. We utilize a transformer module~\cite{vaswani2017attention} with four encoder layers to extract the dynamic features. The transformer module takes the sequence of features $\{f_{\text{cfe}}^i\}_{i=1}^n$ as input and outputs the final features $f \in \mathbb{R}^{D_\text{T}}$.

\paragraph{Training and Inference.}
Following the standard paradigm of camera-based ReID~\cite{chen2020temporal,hou2020temporal,yan2020learning,zhang2020multi}, we exploit a combination of cross-entropy and batch-hard triplet terms~\cite{hermans2017defense} as the loss function, with the hyperparameter $\gamma$:
\begin{equation}
	\mathcal{L} = \mathcal{L}_{\text{ce}} + \gamma \mathcal{L}_{\text{tri}}.
\end{equation}

During inference, the similarity between query and gallery set is measured using the cosine distance.

%% file: sec/3_dataset.tex
\begin{figure*}[]
	\begin{center}
		
		\begin{minipage}{0.5\linewidth}
			\begin{center}
				\includegraphics[width=\linewidth]{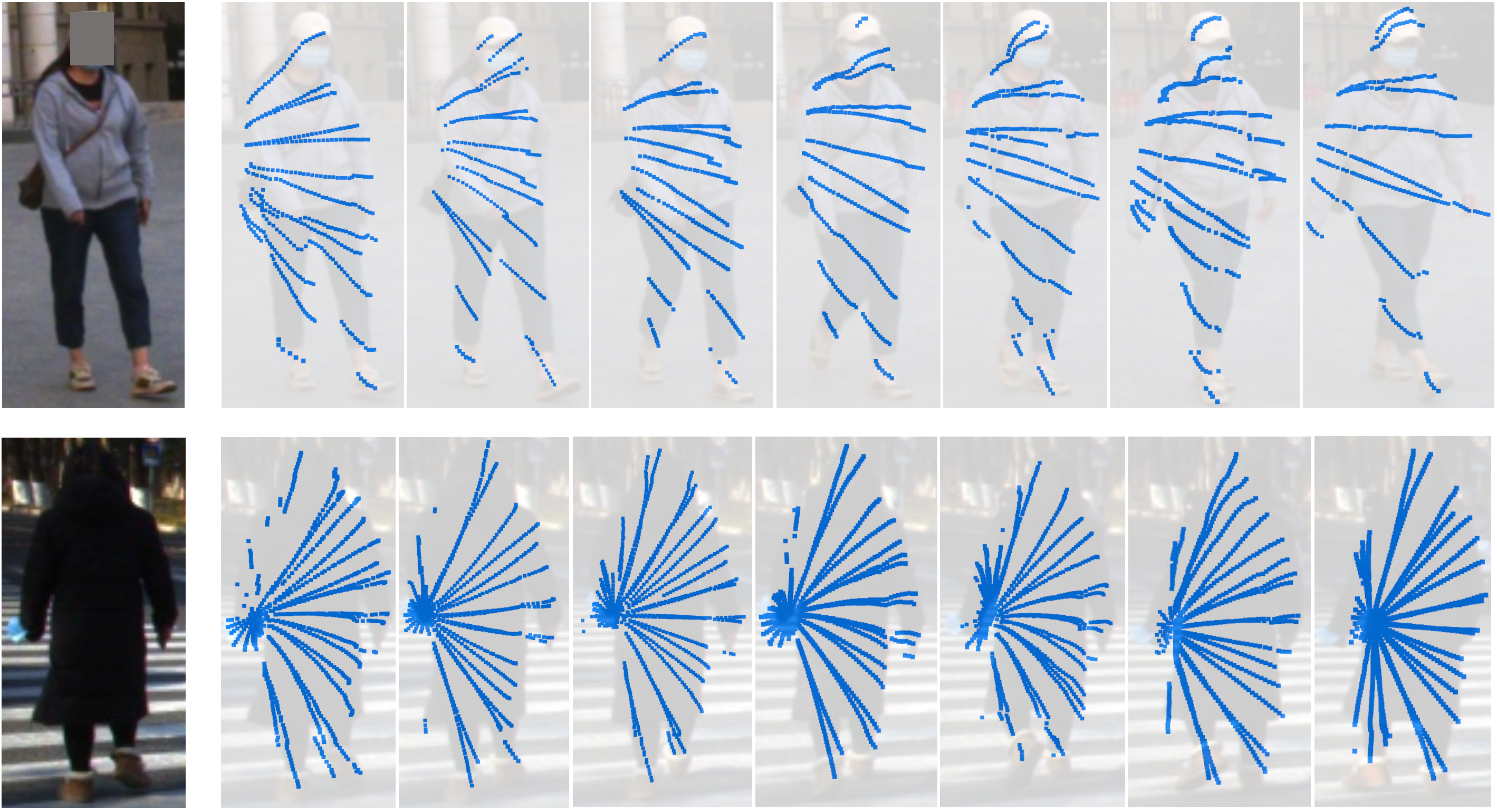}
				\caption{Samples of two pedestrians collected from the square scene in summer (top row), and the crossroads scene in winter (bottom row).}\label{fig:dataset_twoperson}
			\end{center}
		\end{minipage}
		\quad\quad\quad
		\begin{minipage}{0.38\linewidth}
			\begin{center}
				\includegraphics[width=0.945\linewidth]{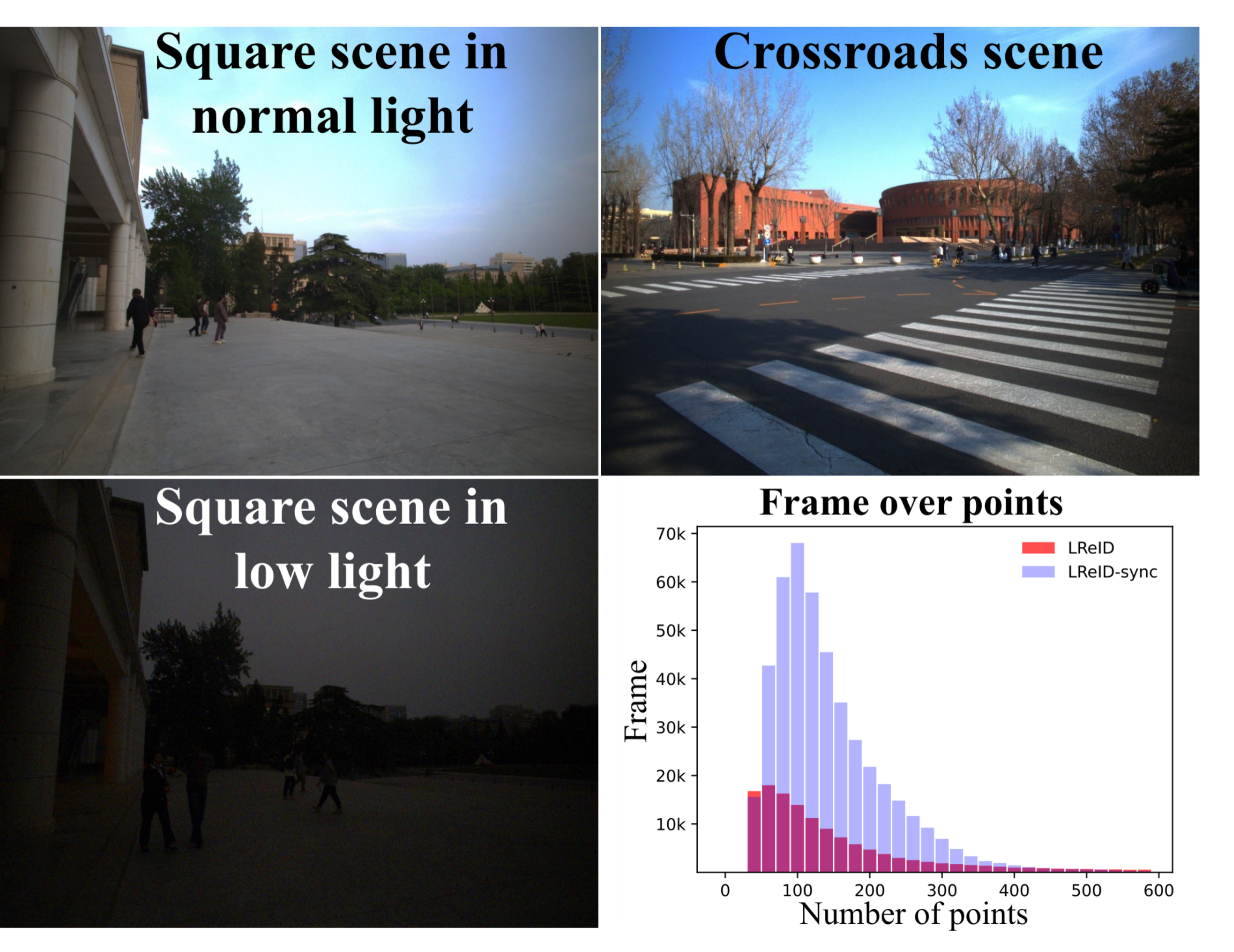}
				\caption{Data acquisition scenes and statistics about points in LReID and LReID-sync (lower right).}
				\label{fig:dataset_scene}
			\end{center}
		\end{minipage}	
	\end{center}
	
\end{figure*}

\section{Person ReID Dataset}
\label{sec:dataset}
In this paper, we build a real-world LiDAR-based ReID dataset LReID and a simulated pedestrian point cloud dataset LReID-sync. LReID is captured by a multimodal imaging system, which consists of 4 Livox Mid-100 LiDAR sensors and 4 industrial cameras. LReID includes 320 identities, 156,000 point cloud frames and synchronized RGB images with corresponding 2D and 3D annotations.

\paragraph{Data Acquisition.}
LReID is collected in two extensive outdoor scenes: a crossroad and a square in front of a building, capturing different time periods and weather conditions, as shown in Fig.~\ref{fig:dataset_scene}. We synchronize each pair of LiDAR and camera with $\sim5$ ms accuracy and provide an accurate extrinsic and intrinsic calibration. All sequences are recorded with a frame rate of 10 Hz. The images are captured of a resolution of 2048 $\times$ 1536 pixels and the density of LiDAR points is about 30,000 points per frame. 

The square scene is located in front of a building with frequent pedestrian traffic, and many pedestrians walk together in groups, which makes it more complex for person ReID tasks. We collected data in the square scene during two different times in summer to analyze the impact of lighting conditions on person ReID. Additionally, to explore the impact of different scenes, weather conditions, and pedestrian attire in different seasons, we also collected data at a crossroads scene in winter. Figure~\ref{fig:dataset_twoperson} shows the samples of two pedestrians collected from different scenes. Furthermore, we analyze the point distribution in LReID and LReID-sync, as shown in Fig.~\ref{fig:dataset_scene}.

\paragraph{Annotations.}
To ensure efficient and accurate annotation, we utilize a point cloud-based 3D detector~\cite{lang2019pointpillars} to locate pedestrians. Additionally, we employ a 3D multi-object tracking method~\cite{weng2020ab3dmot} to establish continuous trajectories, and manually rectify them. Furthermore, we adjust the IDs of the pedestrians that appeared repeatedly in the field of view based on the images. After obtaining precise 3D ReID annotations, we project them onto synchronized RGB images and perform manual corrections.

To avoid the influence of absolute coordinate information on the network, we normalize the pedestrian point clouds by subtracting the center coordinates of their respective bounding boxes.

\paragraph{Evaluation.}
LReID is divided into two splits: a training set with 220 identities and a test set with the remaining 100 identities. Both sets include different scenes, seasons, and lighting conditions. 
Within the test set, 30 identities are captured under low light, whereas 70 identities are captured under normal light. Following the standard paradigm of video-based datasets~\cite{zheng2016mars,wu2018exploit,li2019global}, the evaluation focuses on recognition across different LiDARs.

We select one sample of each identity in the test set to build the query set and use the other samples as the gallery set.
In the test set, we ensure that the data collected by different LiDARs does not overlap in time, which is to prevent the model from recognizing individuals based on human pose at a certain time. 
To maximize the use of training data, we divide the sequences in training set into several fragments that may partially overlap in time. 
The evaluation metrics for LReID include Cumulative Matching Characteristics (CMC) and mean Average Precision (mAP), which are consistent with camera-based datasets.

\begin{figure}
	\centering
	\includegraphics[width=0.97\linewidth]{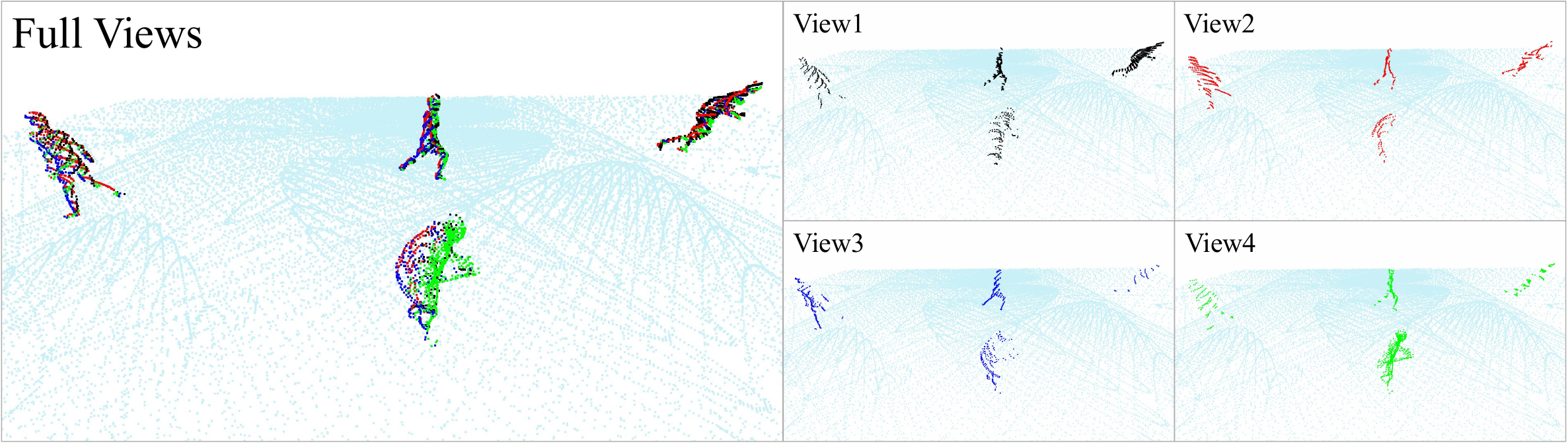}
	\caption{Samples of LReid-sync for full views and each single view.}
	\label{fig:sync}
\end{figure}

\paragraph{LReID-sync.}
LReID-sync is a novel pedestrian dataset generated using the software Unity3D, which simulates pedestrians in scenes captured by multiple synchronized LiDARs from various views, as shown in Fig.~\ref{fig:sync}. With this setup, LReID-sync accurately simulates the information loss and point cloud sparsity that occur when pedestrians are captured using a single-view LiDAR. Additionally, LReID-sync includes annotations for point cloud completion from single view to full views, as well as SMPL parameters. Therefore, LReID-sync can be utilized for pre-training. LReID-sync comprises 600 pedestrians with various actions, each exhibiting unique body shapes and gaits, ensuring diversity within the dataset.

%% file: sec/5_experiment.tex
\begin{table*}[]
	\footnotesize
	\begin{center}
	\caption{Comparison with state-of-the-art camera-based methods on LReID. Numbers in bold indicate the best performance and underscored ones are the second best. To ensure fairness, ReID3D without pre-training is evaluated. }\label{tab:camera}
	\vspace{-.1cm}
	\begin{threeparttable}
	\begin{tabular}{ccc|ccc|ccc|ccc}
		\hline
		\multirow{2}*{Method} & \multirow{2}*{Venue} & \multirow{2}*{Modality} & \multicolumn{3}{c|}{Normal light} & \multicolumn{3}{c|}{Low light} & \multicolumn{3}{c}{Overall} \\ 
		~& ~ & ~ & Rank-1 &Rank-3&mAP&Rank-1&Rank-3&mAP&Rank-1&Rank-3&mAP\\ \hline
		TCLNet~\cite{hou2020temporal}& ECCV 2020 & \multirow{4}*{Camera} & \textbf{98.6} &\underline{98.6}&\textbf{94.88}&60.0&73.3&46.18&87.0&91.0&80.27\\
		STMN~\cite{eom2021video}& ICCV 2021 & ~ & 94.3 &\underline{98.6}&93.81&30.0&50.0&31.17&75.0&84.0&75.02\\
		SINet~\cite{bai2022salient}& CVPR 2022 & ~ & \underline{97.1} &\textbf{100.0}&\underline{94.63}&43.3&60.0&43.97&81.0&88.0&79.43\\ 
		PiT~\cite{zang2022multidirection}& TII 2022 & ~ & 94.3&95.7&86.21&33.3&60.0&35.74&76.0 &85.0&71.07\\ \hline
		B-ReID3D& Ours & \multirow{2}*{LiDAR} & 90.0 & 97.1 &82.04& \underline{90.0} &\underline{93.3}&\underline{81.64}&\underline{90.0} & \underline{96.0} & \underline{81.92} \\
		ReID3D& Ours & ~ & 94.3 & \underline{98.6} &83.65& \textbf{93.3} &\textbf{96.7}&\textbf{82.43}&\textbf{94.0} & \textbf{98.0} & \textbf{83.28}\\ \hline
	\end{tabular}
	\begin{tablenotes}
		\footnotesize 
		\item B-ReID3D: ReID3D without pre-training.
	\end{tablenotes}
	\end{threeparttable} 
	\end{center}
	\vspace{-.2cm}
\end{table*}

\section{Experiments}
\label{sec:experiments}
\subsection{Implementation Details}

\paragraph{ReID Network.}
The ReID network is trained for 700 epochs using the AdamW optimizer with a weight decay of 5e-5. The learning rate, initially set to 5e-5, is updated using a Cosine Annealing Learning Rate (CosineAnnealingLR) scheduler with a cycle of 200 epochs. Each frame of pedestrian point clouds in LReID is upsampled or downsampled to 256 points as input to the network. To train our model, we randomly choose 6 identities, and sample 6 sequences for each identity with a sequence length of 30 frames. The loss weight $\gamma$ is set to 1.
The neighborhood size is set to 10 and 8 for each graph convolution layer and Binarization module, respectively. And the feature dimensions $D$, $D_\text{C}$ and $D_\text{T}$ are set to 512, 512 and 1024, respectively.

\paragraph{Pre-training.}
The pre-training model follows a similar training strategy as described above, except that the initial learning rate is set to 1e-4. The loss weight $\eta$ is set to 1. The coefficient $\delta$ in Eq.~\ref{eq:a} is set to 0.01 for the first 100 epochs, then increased to 0.1, 0.5 and 1.0 after 100, 200 and 400 epochs. 
The coarse shape $\hat{\mathcal{P}}_{\text{coarse}}$ contains 128 points, while the detailed shape $\hat{\mathcal{P}}_{\text{detail}}$ contains 512 points. The weights of the pre-trained GCEE are used as initialization for the ReID network.

\subsection{Comparison with Camera-based Methods}
We compare ReID3D with the state-of-the-art video-based methods~\cite{eom2021video,hou2020temporal,zang2022multidirection,bai2022salient}. The input images are resized in the resolution of 128 $\times$ 64. To ensure fairness, ReID3D without pre-training (B-ReID3D) is also evaluated. All methods are evaluated with the same dataset settings and metrics. 

The comparative results are shown in Tab.~\ref{tab:camera}, from which the following observations can be obtained: (1) ReID3D and B-ReID3D demonstrate their superiority to the video-based methods, primarily benefiting from the utilization of point clouds, which is unaffected by lighting conditions and complex background. (2) ReID3D achieves state-of-the-art results in overall and low light conditions, but it falls behind video-based methods in normal light conditions. This is because video-based methods make full use of appearance information under normal light. (3) Video-based methods perform poorly under low light, while ReID3D and B-ReID3D demonstrate comparable reliability under both low light and normal light.

\subsection{Pre-training}
\begin{table}
	\footnotesize
	\centering
	\caption{Comparison for different pre-training method.}\label{tab:transfer}
	\vspace{-.1cm}
	\begin{tabular}{c|ccc}
		\hline
		Pre-training Method & Rank-1 & Rank-3 & mAP\\ \hline
		Baseline(w/o pre-training)& 90.0 & 96.0 & 81.92\\
		ReID& 91.0 & 96.0 & 82.13 \\ 
		Completion& 93.0 & 97.0 & 82.73 \\ 
		Completion + SMPL& \textbf{94.0} & \textbf{98.0} & \textbf{83.28} \\ \hline
	\end{tabular}
\end{table}
To demonstrate the effectiveness of pre-training with simulated dataset LReID-sync, we evaluate the performance of different pre-training method. The following four method are evaluated: (1) ReID3D without pre-training. (2) Pre-training with similar ReID task, in which the pre-training model and loss are consistent with the ReID network. (3) Pre-training with only the branch of point cloud completion. (4) Pre-training with multiple tasks, as described in Section~\ref{sec:pre}.
The experimental results are shown in Tab.~\ref{tab:transfer}. It can be observed that pre-training with multiple tasks makes an improvement to the accuracy of ReID3D, with the rank-1 accuracy increasing from 90.0 to 94.0. This improvement can be attributed to two factors: (1) The simulated data comprises a diverse collection of pedestrian point clouds with various body shapes, which supplements the real-world data. (2) Multi-task pre-training effectively leverages the performance of the encoder. Besides, pre-training with task of ReID or point cloud completion also leads to an improvement in performance.

\begin{figure}	
	\begin{center}
		
		\begin{subfigure}{0.45\linewidth}
			\begin{center}
				\includegraphics[width=\linewidth,height=1.11in]{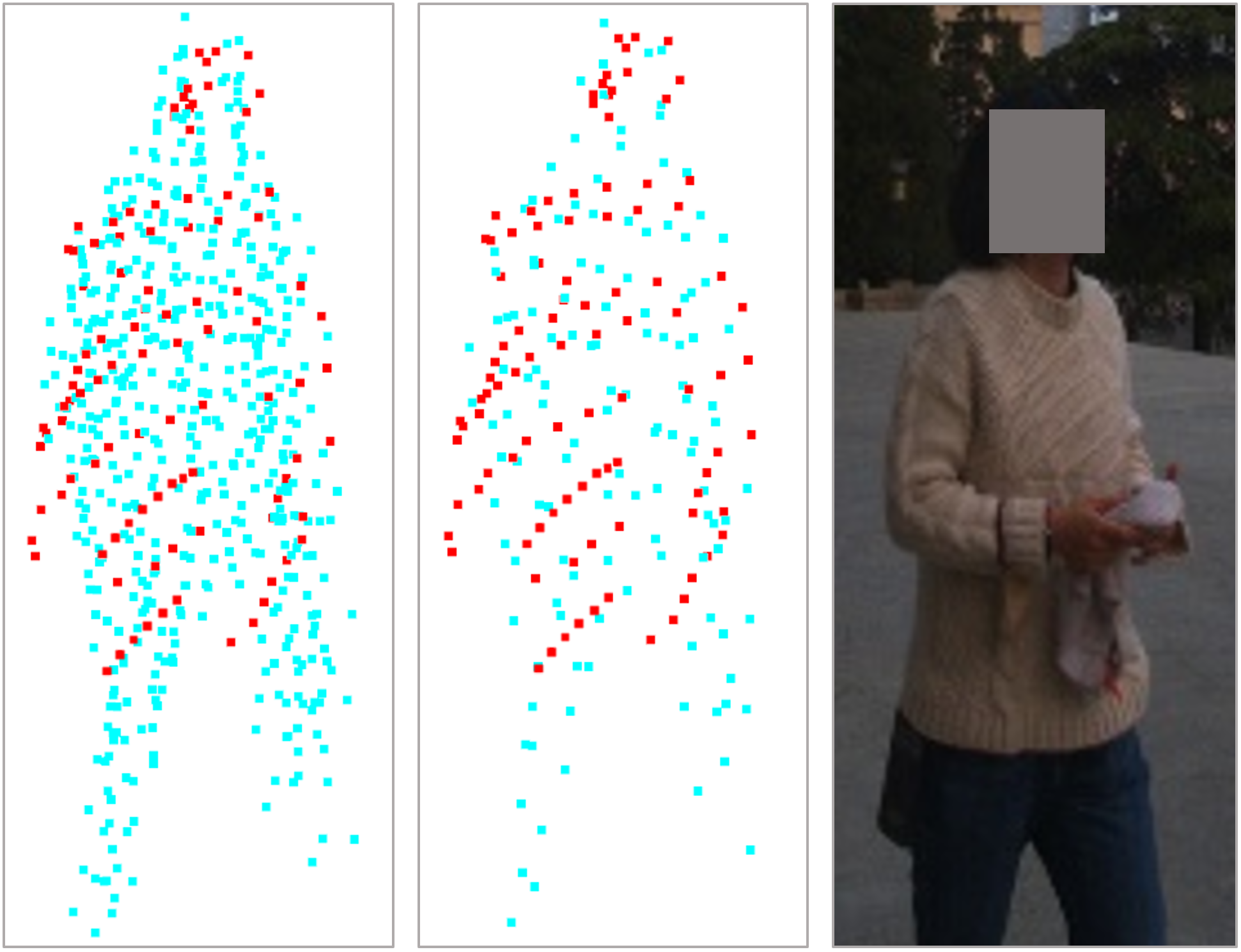}
				\caption{}\label{fig:a}
			\end{center}
		\end{subfigure}
		\quad
		\begin{subfigure}{0.45\linewidth}
			\begin{center}
				\includegraphics[width=\linewidth]{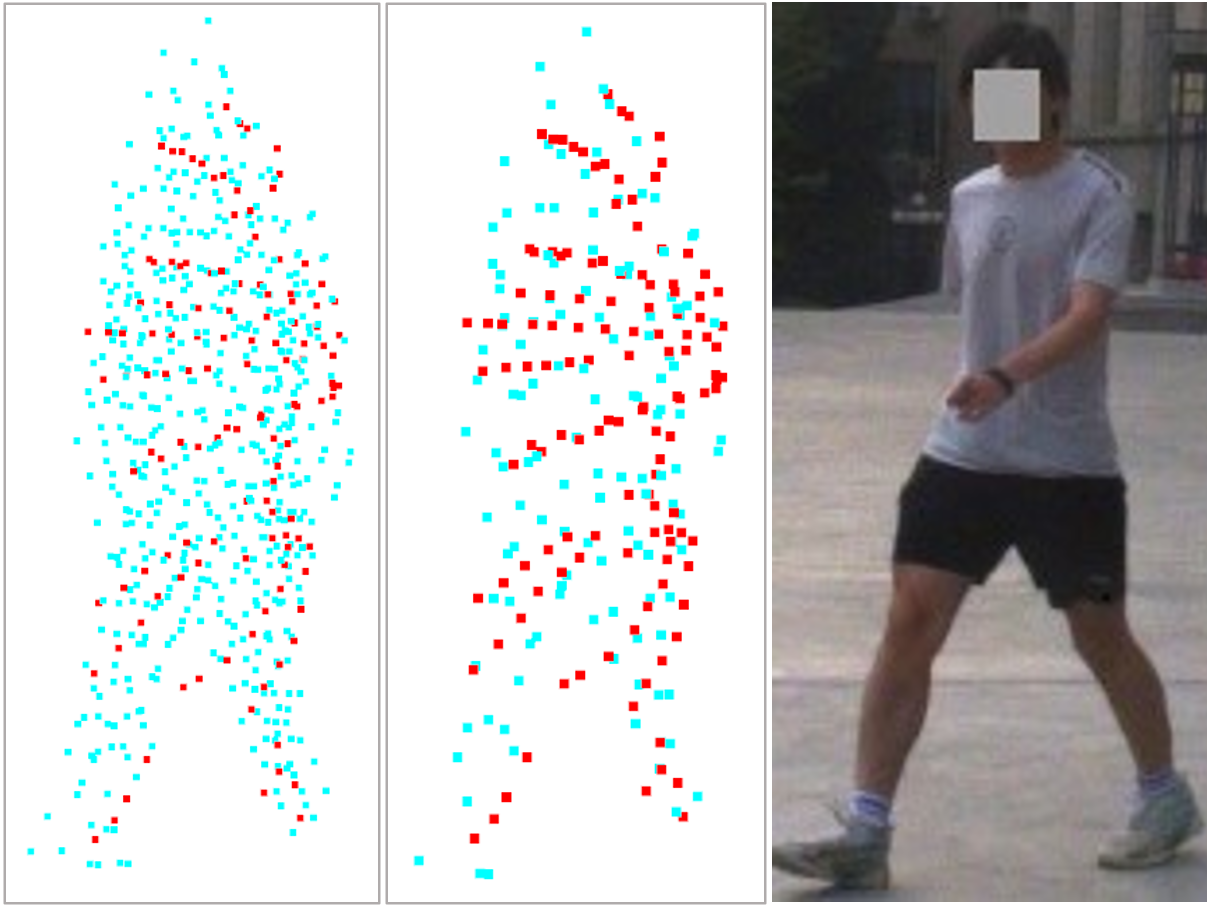}
				\caption{}
				\label{fig:b}
			\end{center}
		\end{subfigure}
		\vskip 0.1cm
		\begin{subfigure}{0.45\linewidth}
			\begin{center}
				\includegraphics[width=\linewidth]{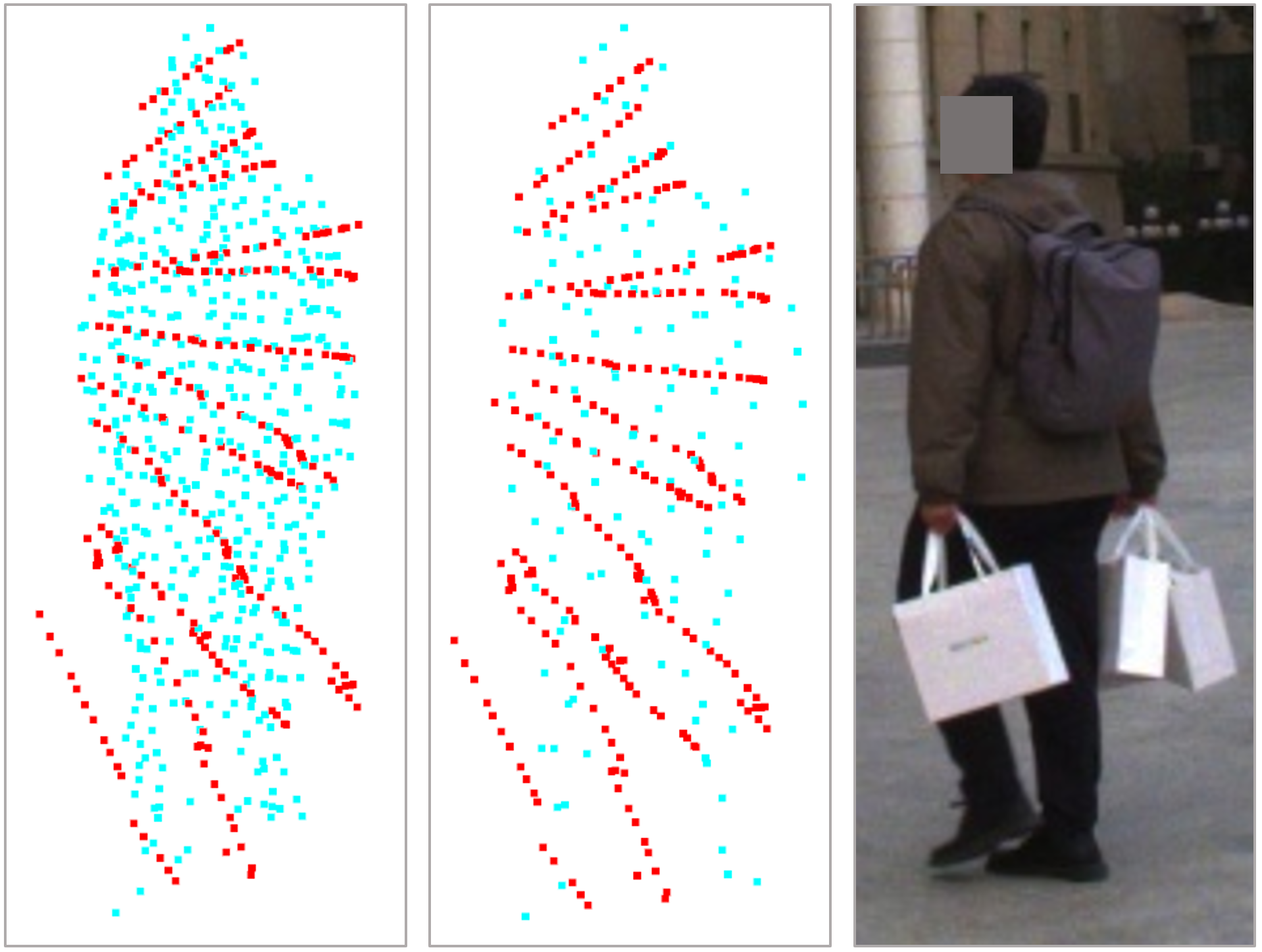}
				\caption{}
				\label{fig:c}
			\end{center}
		\end{subfigure}
		\quad
		\begin{subfigure}{0.45\linewidth}
			\begin{center}
				\includegraphics[width=\linewidth,height=1.11in]{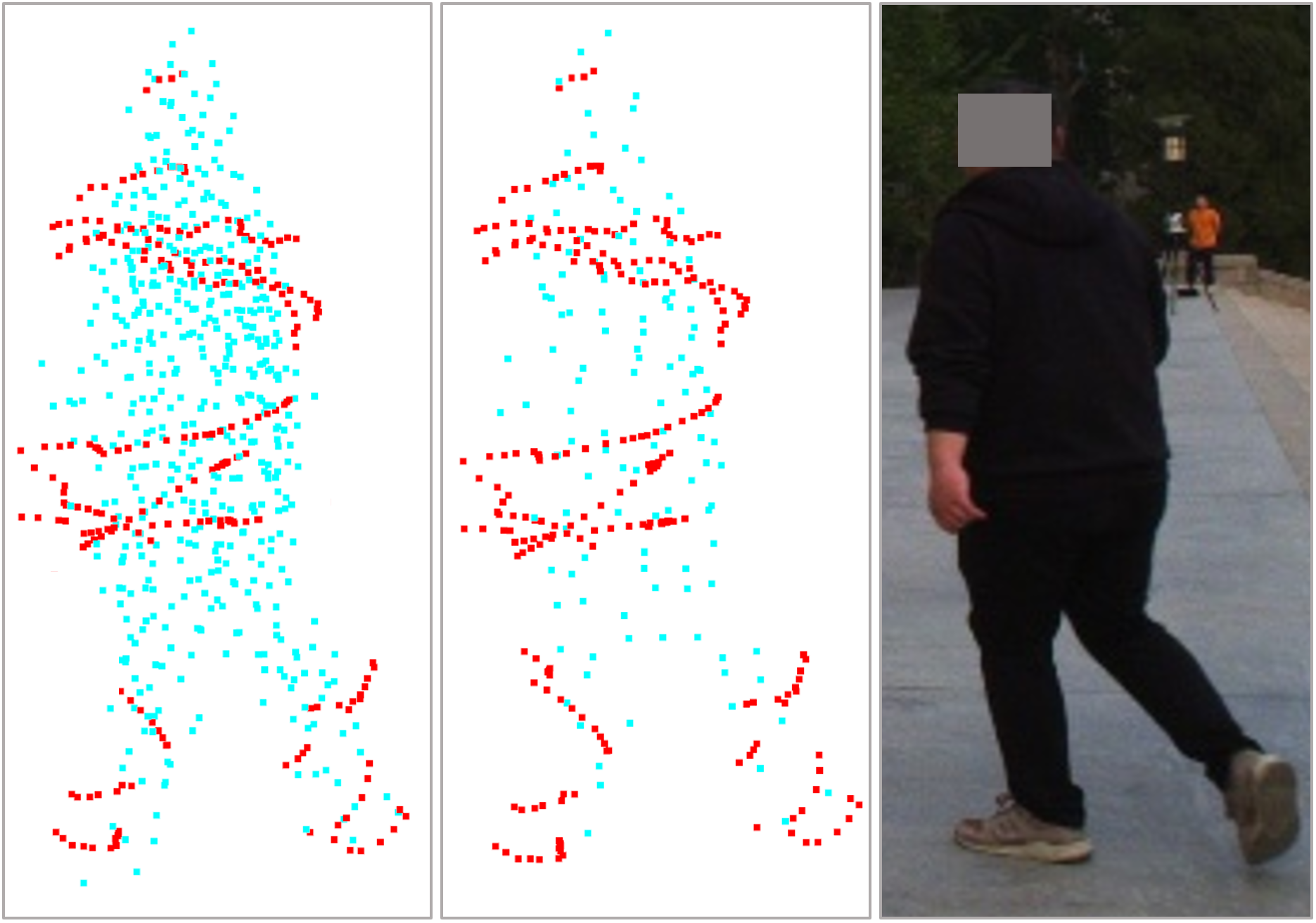}
				\caption{}
				\label{fig:d}
			\end{center}
		\end{subfigure}
	\end{center}
	\vspace{-.2cm}
	\caption{Completion results for real-world pedestrian point clouds obtained by the pre-trained model. In each sample, the detailed shape, coarse shape, and image are shown from left to right. The red points represent the input point cloud, and the blue points represent the predicted complete point cloud.}\label{fig:com}
\end{figure}

To showcase the robust feature encoding capabilities attained by the encoder through pre-training, we visualize the completion results for several real-world pedestrian point clouds with different characteristics, as shown in Fig.~\ref{fig:com}. We can observe that: (1) Intuitively, the detailed and coarse shape closely resemble the actual human shape, which indicates that the encoder has successfully captured the complete features of the human body. (2) The detailed shape is an extension based on the coarse shape, with higher resolution and more information. (3) The pre-trained encoder has the ability to estimate the features of the missing part in the point cloud. As shown in Fig.~\ref{fig:a}, the unobserved parts of the legs are accurately completed. (4) Due to motion blur or object carrying, the point cloud may contain some noise points. For example, in Fig.~\ref{fig:b} and Fig.~\ref{fig:c}, there are red noise points located between the legs and from the carrying bags, respectively. However, the model can effectively filter out the noise points that do not conform to the body shape.

\subsection{Ablation Study}

\paragraph{Effect of Encoders.}
To evaluate the effectiveness of our GCEE, we also implement several commonly used encoders in point cloud tasks to replace GCEE in the ReID network for comparison, including PointNet~\cite{qi2017pointnet}, Point Transformer~\cite{zhao2021point}, and 3DCNN~\cite{choy20194d}. All methods use consistent training strategies and employ a same transformer module as the temporal fusion network. We report the results on LReID in Tab.~\ref{tab:encoder}, obtaining the following observations: (1) GCEE outperforms other encoders, which is mainly beneficial by its flexible graph structure in the feature space and the strategy of complementary enhancement. (2) Point-based encoders are superior to voxel-based encoder 3DCNN, which indicates that point-based encoders can provide a more detailed understanding of humans in tasks like ReID.

\begin{table}
	\footnotesize
	\centering
	\caption{Performance of B-ReID3D with different encoders.}\label{tab:encoder}
	\vspace{-.1cm}
	\begin{tabular}{c|ccc}
		\hline
		Encoder & Rank-1 & Rank-3 & mAP\\ \hline
		PointNet~\cite{qi2017pointnet}& 74.0 & 88.0 & 60.31\\
		Point Transformer~\cite{zhao2021point}& 83.0 & 93.0 & 74.86 \\
		3DCNN~\cite{choy20194d}& 34.0 & 51.0 & 28.27 \\ 		
		GCEE & \textbf{90.0} & \textbf{96.0} & \textbf{81.92} \\ \hline
	\end{tabular}
\end{table}

\paragraph{Effect of CFE.}
\begin{table}
	\footnotesize
	\centering
	\caption{Ablation study on the CFE module and Eraser module.}\label{tab:afe}
	\begin{tabular}{c|ccc}
		\hline
		Encoder Type & Rank-1 & Rank-3 & mAP\\ \hline		
		B-ReID3D-w/o CFE& 83.0 & 94.0 & 74.15 \\ 
		B-ReID3D-w/o Eraser& 85.0 & 95.0 & 80.47 \\ 
		B-ReID3D & \textbf{90.0} & \textbf{96.0} & \textbf{81.92} \\ \hline
	\end{tabular}
\end{table}

CFE is a crucial component in our proposed encoder, used to extract complementary features. We compared our full model with two ablated versions: one without CFE and another without the Eraser module, as shown in Tab.~\ref{tab:afe}. Removing CFE results in a lower rank-1 accuracy, with a score of 83.0 compared to our 90.0, which highlights the effectiveness of the complementary enhancement in constructing distinctive and comprehensive features.
While removing the Eraser module retains the features of the complementary frame, it might lead to the extraction of similar features from the primary frame, thus restricting the improvement in accuracy, with a
score of 85.0 compared to 83.0 of the version without CFE.

\paragraph{Effect of the Temporal Module.}
\begin{table}
	\footnotesize
	\centering
	\caption{Performance of B-ReID3D with different temporal modules.}\label{tab:tem}
	\begin{tabular}{c|ccc}
		\hline          
		Temporal Module & Rank-1 & Rank-3 & mAP\\ \hline		
		LSTM~\cite{gers2000learning}& 82.0 & 89.0 & 72.34 \\ 
		Bi-LSTM~\cite{schuster1997bidirectional}& 80.0 & 90.0 & 70.92 \\	 
		Transformer~\cite{vaswani2017attention} & \textbf{90.0} & \textbf{96.0} & \textbf{81.92} \\ \hline
	\end{tabular}
\end{table}

We evaluate the effectiveness of different temporal modules in combination with GCEE, as shown in Tab.~\ref{tab:tem}. We can observe that the attention-based transformer structure demonstrates significant superiority over the LSTM structure. The rank-1 accuracy of transformer exceeds that of LSTM by 8 points, and that of BiLSTM by 10 points.

\subsection{Visualization Analysis}

\begin{figure}[]
	\begin{center}
		
		\begin{subfigure}{0.31\linewidth}
			\begin{center}
				\includegraphics[height=1.in]{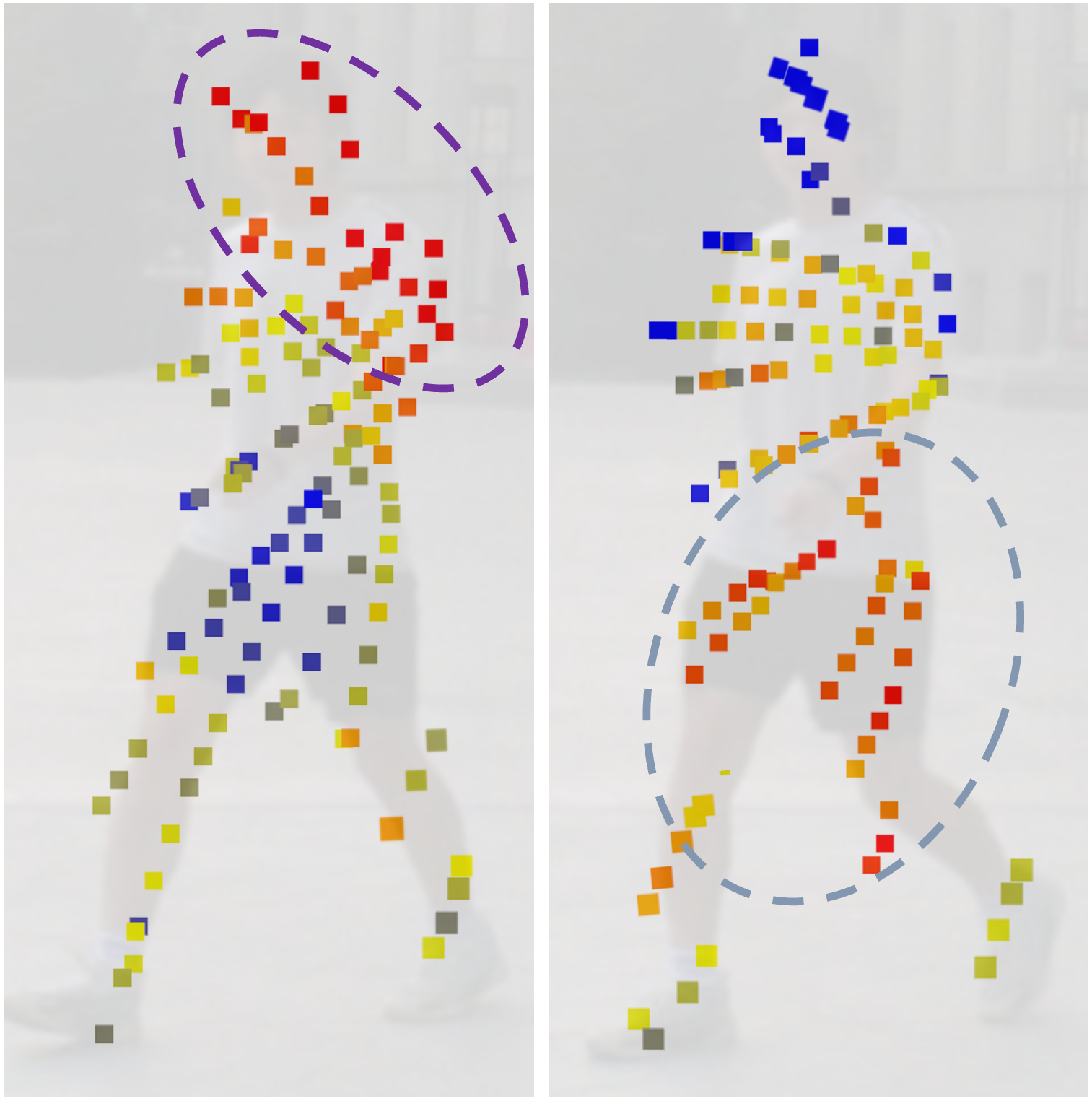}
				\caption{}\label{fig:fea2}
			\end{center}
		\end{subfigure}
		\ \
		\begin{subfigure}{0.27\linewidth}
			\begin{center}
				\includegraphics[height=1.in]{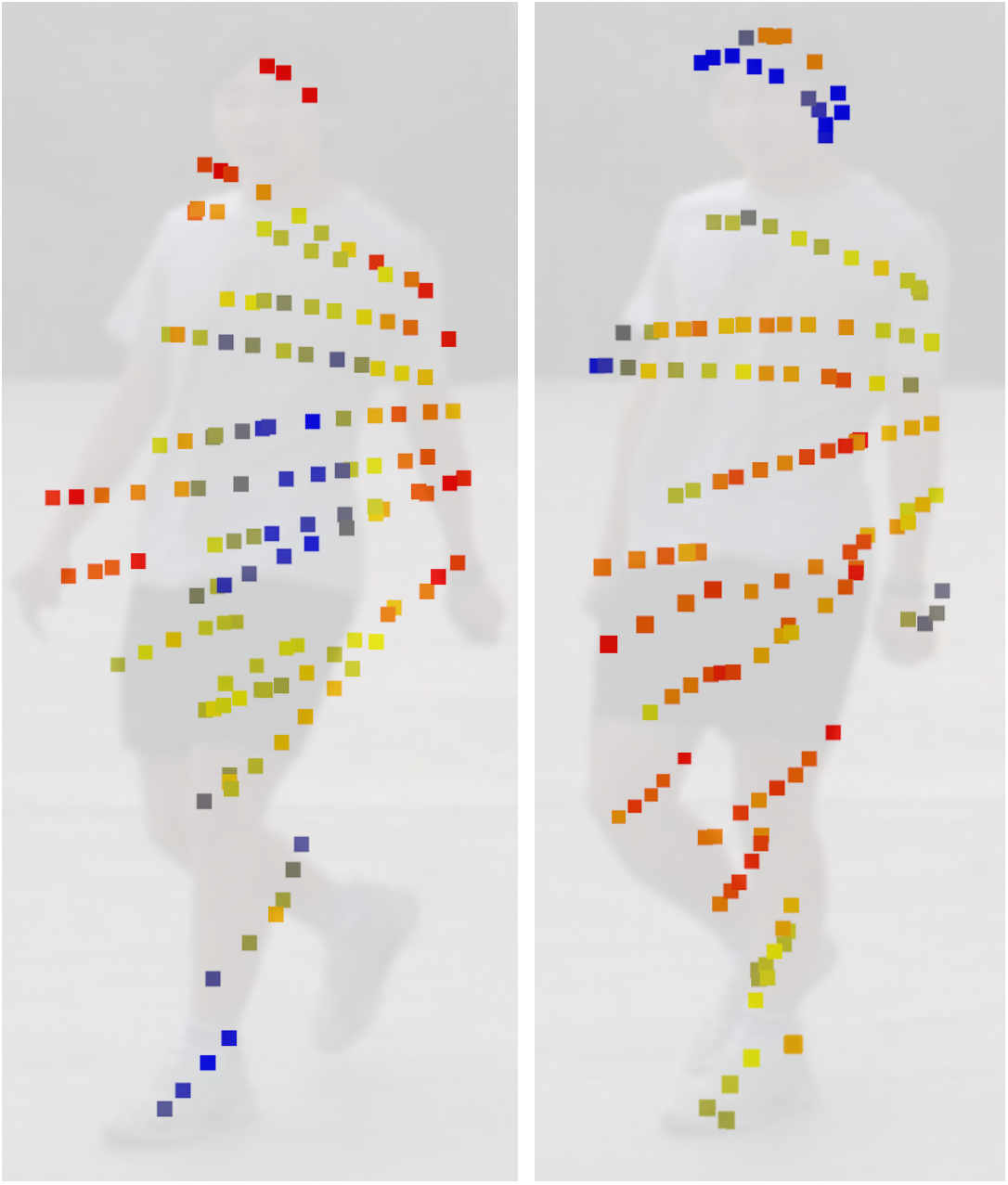}
				\caption{}
				\label{fig:fea2}
			\end{center}
		\end{subfigure}
		\ \
		\begin{subfigure}{0.28\linewidth}
			\begin{center}
				\includegraphics[height=1.in]{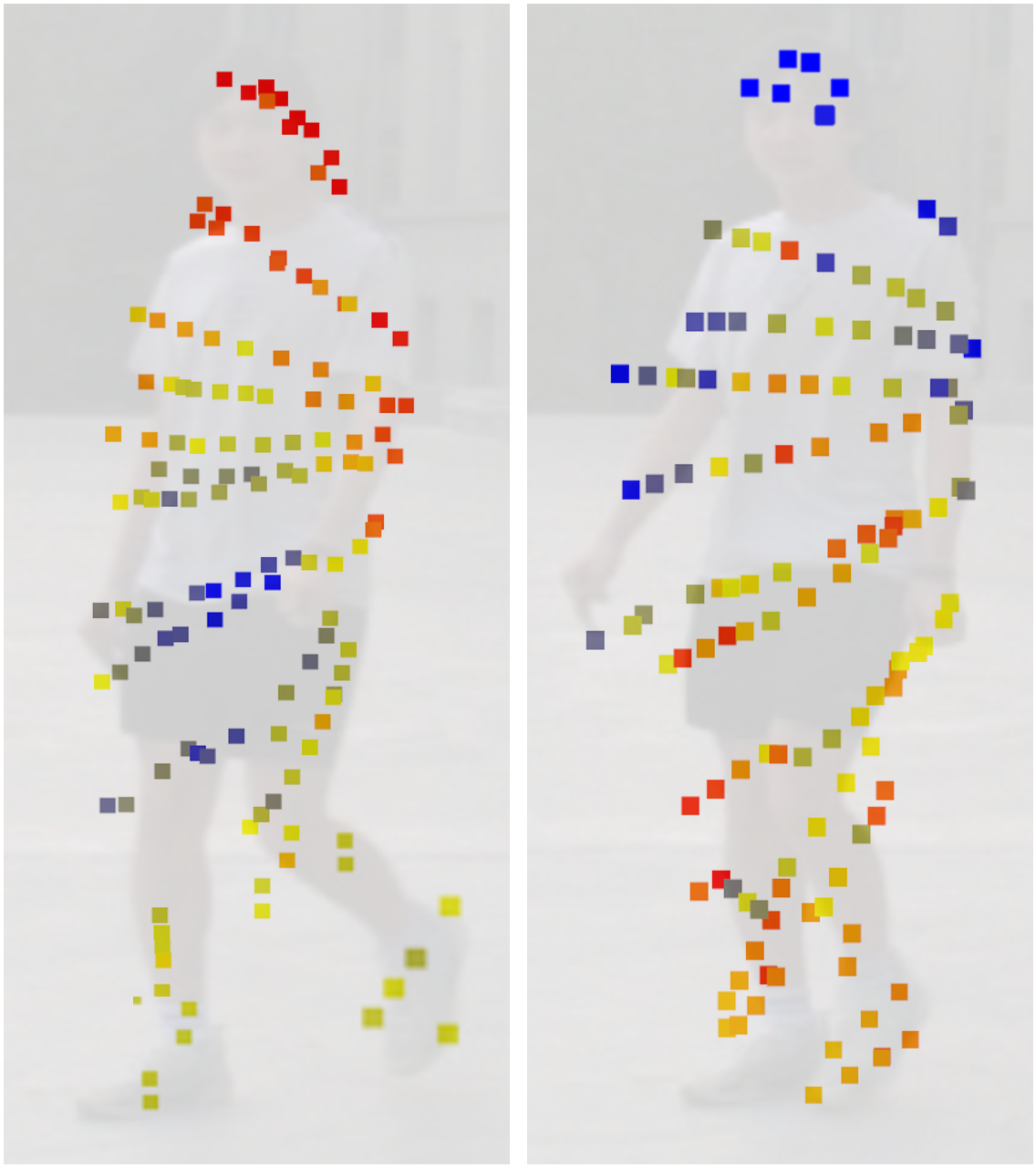}
				\caption{}
				\label{fig:fea3}
			\end{center}
		\end{subfigure}
		\begin{subfigure}{0.03\linewidth}
			\begin{center}
				\includegraphics[height=1.in]{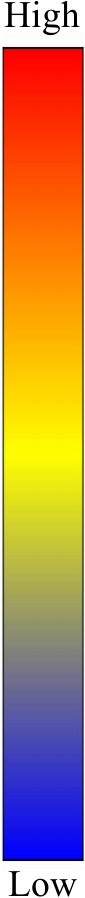}
				\label{fig:tuli}
			\end{center}
		\end{subfigure}
	\end{center}	
	\vspace{-.4cm}
	\caption{Feature visualization of CFE. The features of the primary frame and the supplementary frame are presented on left and right, respectively. The color bar on the right indicates the feature value.}\label{fig:fea}
	
\end{figure}

For qualitative analysis, we visualize the feature values output by the two graph convolutional layers in the CFE, as shown in Fig.~\ref{fig:fea}. 
Three samples are taken from different actions of a person in the test set of LReID, with each sample comprising two consecutive frames. In each sample, the features of the primary frame and the supplementary frame are presented. 
We can observe that primary ones pay more attention to the head and arms, potentially linked to the discriminative height parameter, while supplementary ones predominantly emphasize the torso and legs. The combination of them creates a complementary and comprehensive feature representation.

%% file: sec/6_conclu.tex
\section{Conclusions}
\label{sec:conclu}

This paper presents the first research on person ReID using precise 3D structural information provided by LiDAR. 
Firstly, We propose a LiDAR-based ReID framework, named ReID3D, that utilizes pre-training to guide Graph-based Complementary Enhancement Encoder (GCEE) for extracting comprehensive 3D intrinsic features. 
Moreover, we build the first LiDAR-based person ReID dataset, termed LReID, which contains 320 pedestrians in various outdoor scenes and lighting conditions. 
Additionally, we introduce LReID-sync, a new simulated pedestrian dataset designed for pre-training encoders with tasks of point cloud completion and shape parameter learning. 
Our proposed ReID3D demonstrates exceptional performance on LReID, highlighting the significant potential of LiDAR in addressing person ReID tasks. 

%% file: sec/X_suppl.tex
\clearpage
\setcounter{page}{1}
\appendix
\maketitlesupplementary

\section{LReID}
We provide detailed instructions about our data acquisition system and scenes to supplement the data collection process of LReID. 

\subsection{Data Acquisition System}
\begin{figure}
	\begin{center}
		\includegraphics[width=0.95\linewidth]{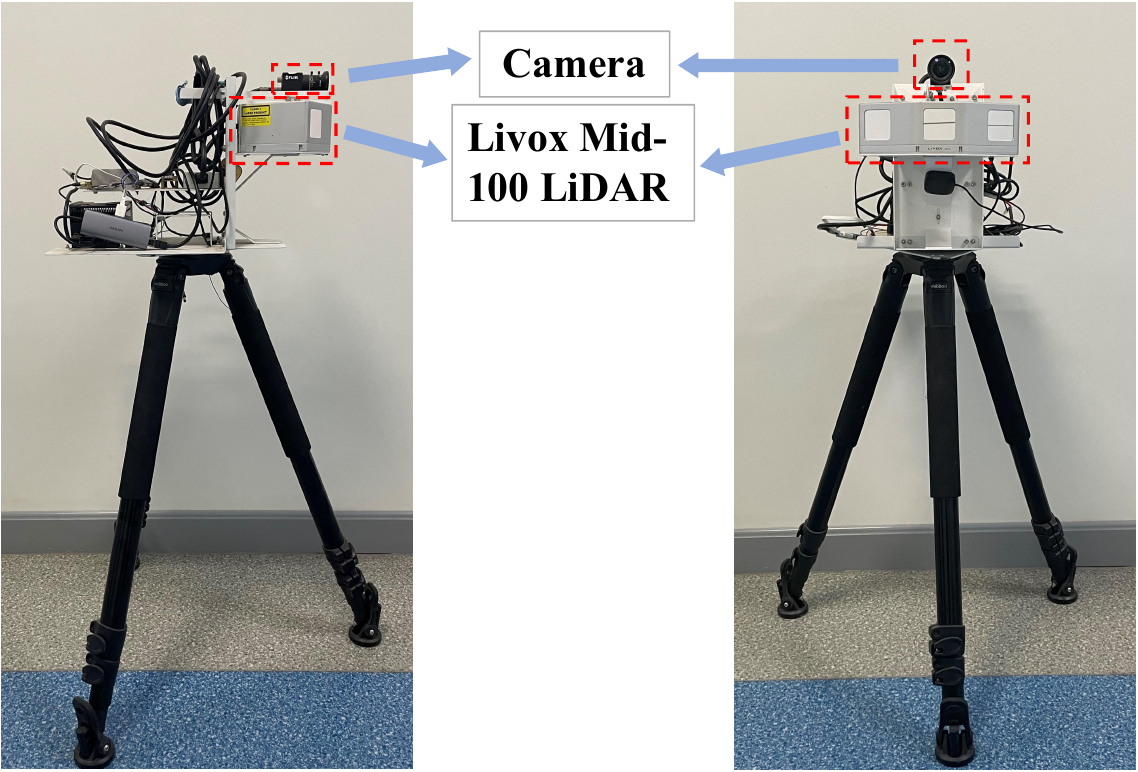}
	\end{center}
	\caption{The acquisition node.}
	\label{fig:mid100}
\end{figure}

We collect LReID with 4 acquisition nodes, each including a Livox Mid-100 LiDAR and an industrial camera, as shown in Fig.~\ref{fig:mid100}. The acquisition nodes are portable and powered by their carried power supplies. The Livox Mid-100 LiDAR continuously scans the scene at about 30,000 points per frame for a horizontal view of 100° and a vertical view of 40° toward the front, providing accurate depth information. The Livox Mid-100 LiDAR is a low-cost option, priced at around \$1,400 \footnote{\href{https://store.dji.com/product/livox-mid-100-5pcs?vid=49181&set_region=US&from=store-nav}{Purchase link}}, which expands the practical applications of LiDAR-based person ReID. 

\subsection{Data Acquisition Scenes}
\begin{figure*}[]
	\begin{center}
		
		\begin{subfigure}{0.9\linewidth}
			\begin{center}
				\includegraphics[width=\linewidth]{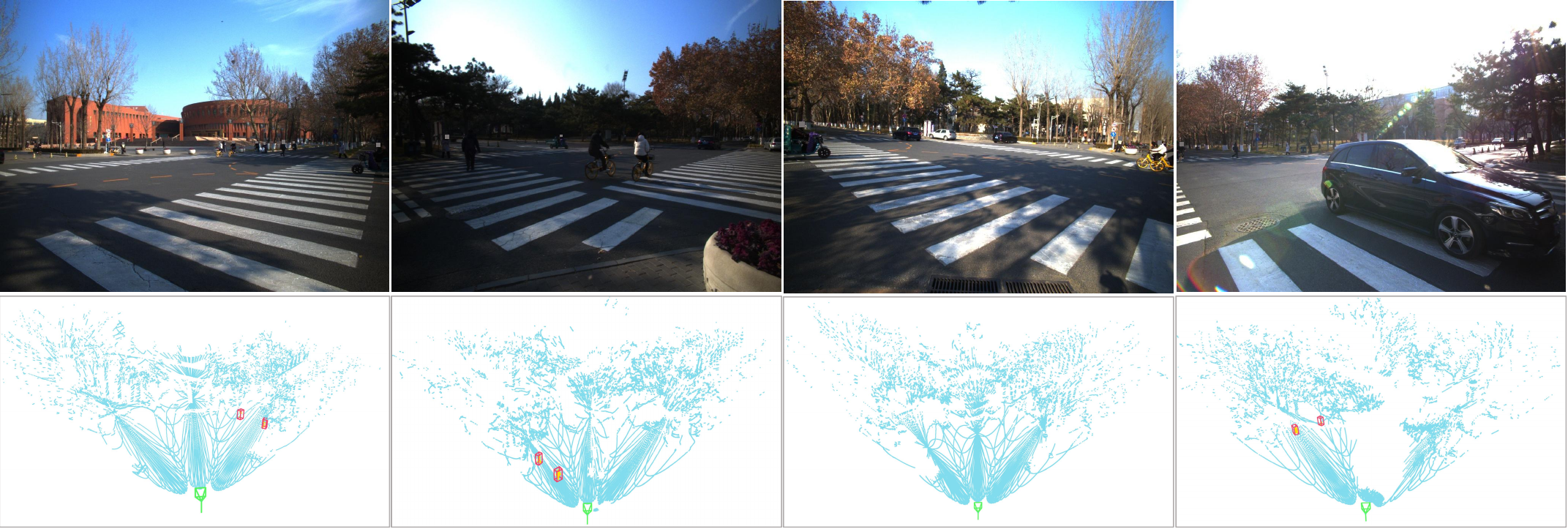}
				\caption{Crossroads scene.}\label{fig:crossroads}
			\end{center}
		\end{subfigure} \\ \vspace{.3cm}
		\begin{subfigure}{0.9\linewidth}
			\begin{center}
				\includegraphics[width=\linewidth]{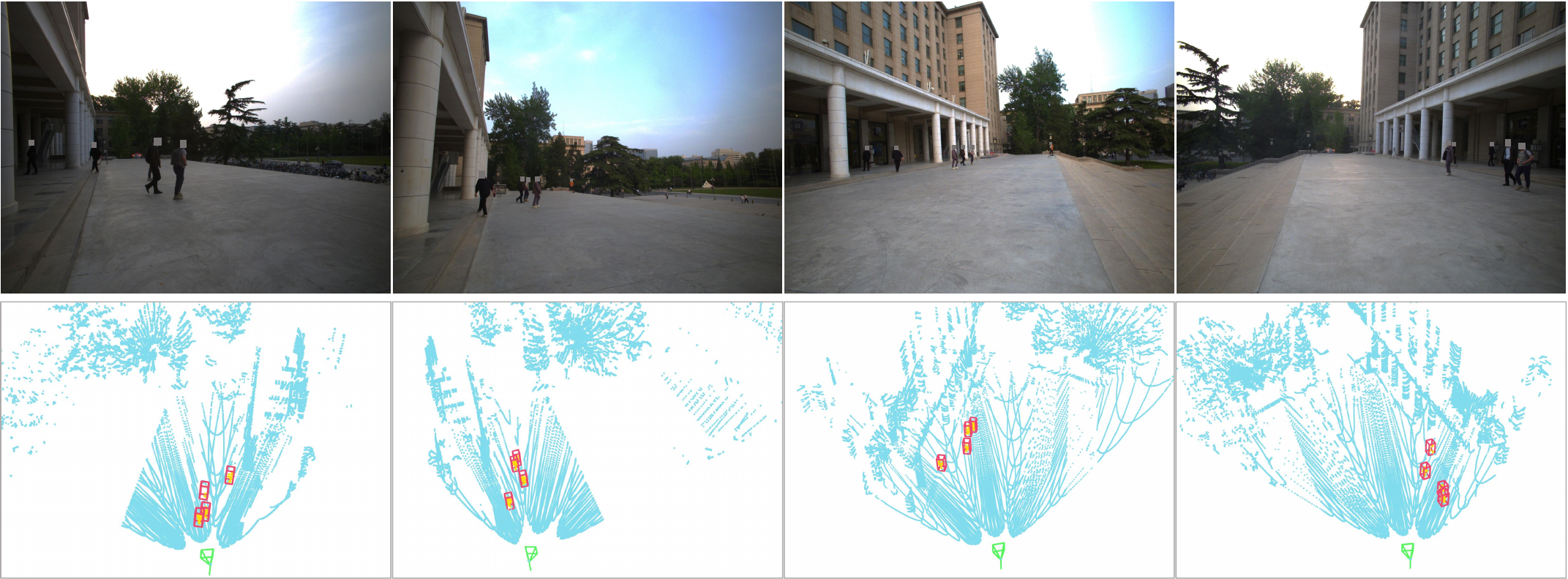}
				\caption{Square scene in normal light condition.}
				\label{fig:zhulou1}
			\end{center}
		\end{subfigure}\\ \vspace{.3cm}
		\begin{subfigure}{0.9\linewidth}
			\begin{center}
				\includegraphics[width=\linewidth]{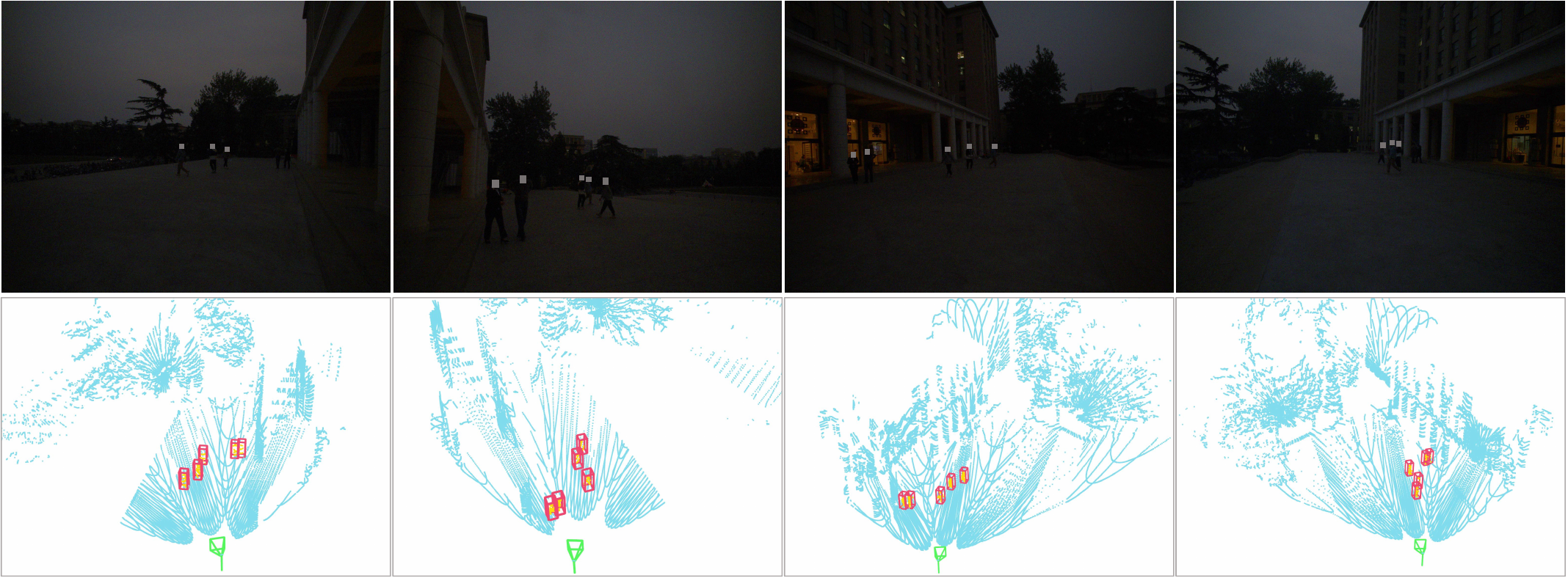}
				\caption{Square scene in low light condition.}
				\label{fig:zhulou2}
			\end{center}
		\end{subfigure}
	\end{center}	
	\vspace{-.3cm}
	\caption{Data acquisition samples in different scenes and lighting conditions.}\label{fig:scenes}
	
\end{figure*}

With the aim of improving the generalization and robustness of models trained on LReID, as well as expanding the data coverage, we collect LReID in various outdoor scenes: a crossroad and a square, capturing different time periods and weather conditions, as shown in Fig.~\ref{fig:scenes}.

\section{Feature Visualization}
To visually demonstrate the superiority of our proposed ReID3D, we use t-SNE to visualize the feature distributions of LReID test set extracted by different methods. 

\subsection{Comparison with Camera-based Method}
\begin{figure*}
	\begin{center}
	\includegraphics[width=0.99\linewidth]{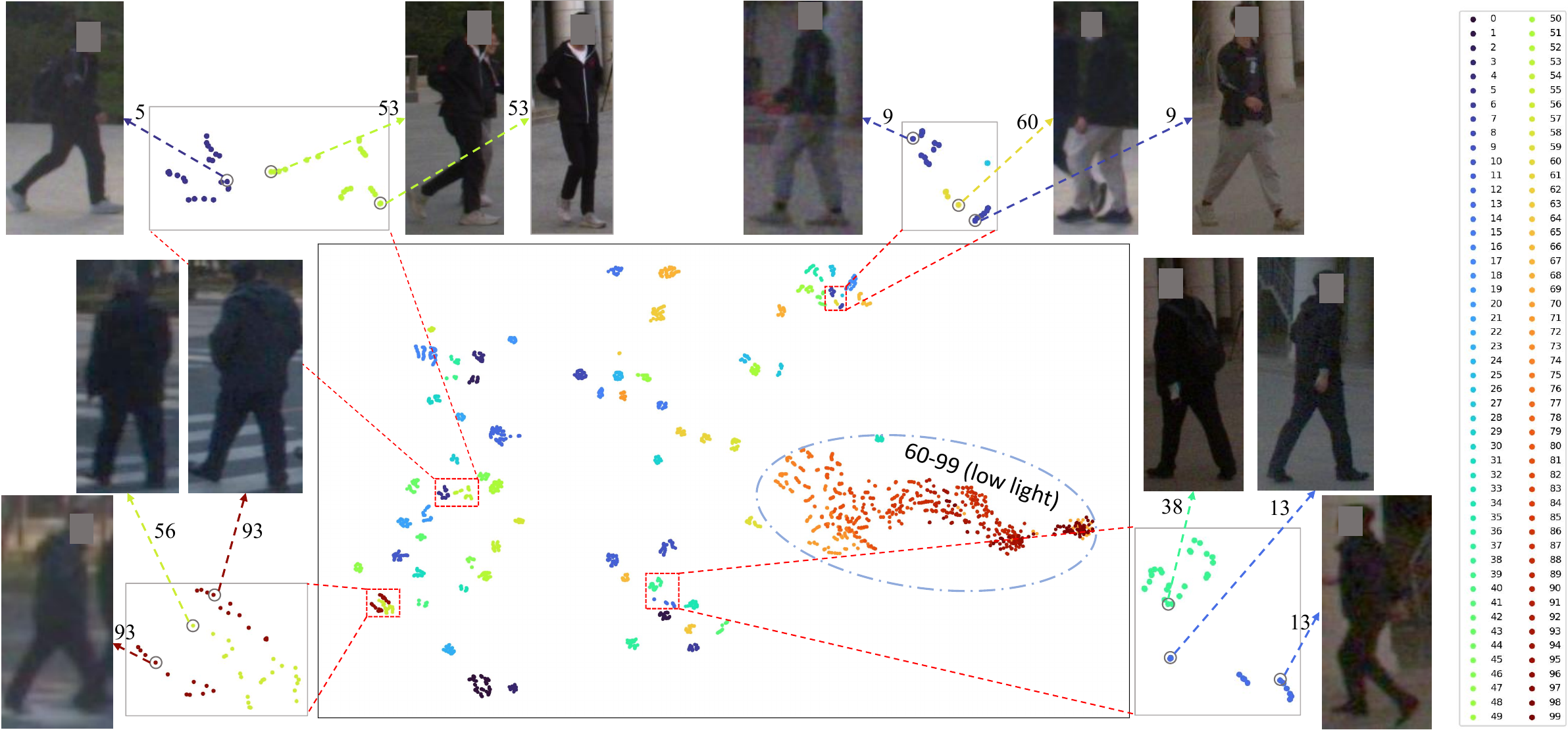}
	\end{center}
	\caption{The feature distribution of LReID test set extracted by TCLNet~\cite{hou2020temporal}. Each point corresponds to a feature vector output by TCLNet and distinct colors represent different individuals in LReID test set. Each image is selected from the corresponding sample of the pedestrian. It is challenging to distinguish the features of pedestrians with similar appearances and those under low light.}
	\label{fig:tsne_tcl}
\end{figure*}

For the purpose of comparing the disparities  between different modalities, we visualize the feature distributions obtained from TCLNet~\cite{hou2020temporal} that achieves the highest accuracy among camera-based methods and ReID3D without pre-training (B-ReID3D), as shown in Fig.~\ref{fig:tsne_tcl} and Fig.~\ref{fig:tsne_reid3d}. We can observe that: (1) It is difficult to distinguish pedestrians in low light condition for TCLNet, with their features entangled in the feature space. However, their features extracted by B-ReID3D are dispersed in the feature space, like the features of pedestrians under normal light. (2) TCLNet is prone to confusion between individuals with similar appearances, as illustrated by the four examples in Fig.~\ref{fig:tsne_tcl}. In contrast, their features extracted by B-ReID3D display excellent discriminability. This is because camera-based methods primarily focus on learning the visual appearance of pedestrians, while ReID3D is able to acquire intrinsic features, such as body shape and gait. (3) B-ReID3D also exhibits mistakes in distinguishing certain pedestrians, which are easily discernible for camera-based methods. Therefore, methods based on the two modalities have complementarities.

\subsection{Effect of Pre-training}
\begin{figure*}
	\begin{center}
		\includegraphics[width=0.9\linewidth]{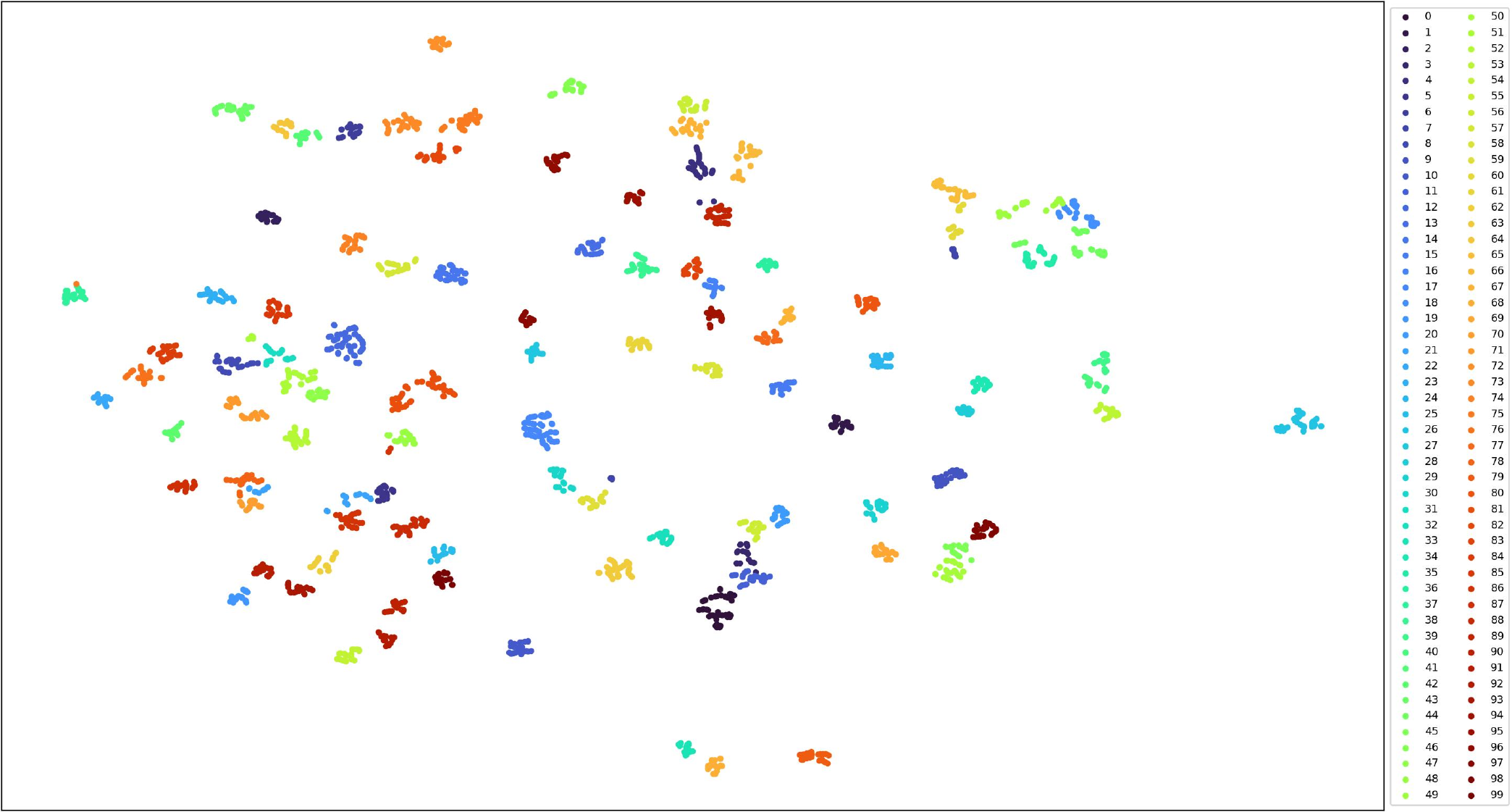}
	\end{center}
	\caption{The feature distribution of LReID test set extracted by ReID3D.}
	\label{fig:tsne_pre}
\end{figure*}

In order to demonstrate the effectiveness of pre-training with multiple tasks, we visualize the feature distribution obtained from ReID3D, as shown in Fig.~\ref{fig:tsne_pre}. Compared to Fig.~\ref{fig:tsne_reid3d}, the feature distributions extracted by ReID3D and B-ReID3D exhibit overall similarity. However, ReID3D demonstrates better discriminative ability when dealing with individuals with similar features in local regions of the feature space. We attribute this to the fact that LReID-sync improves the diversity of the actual training data, and the pre-training with multiple tasks enhances GCEE in learning features of human body shape.

\subsection{Comparison with Point Transformer}
\begin{figure*}
	\begin{center}
	\includegraphics[width=0.99\linewidth]{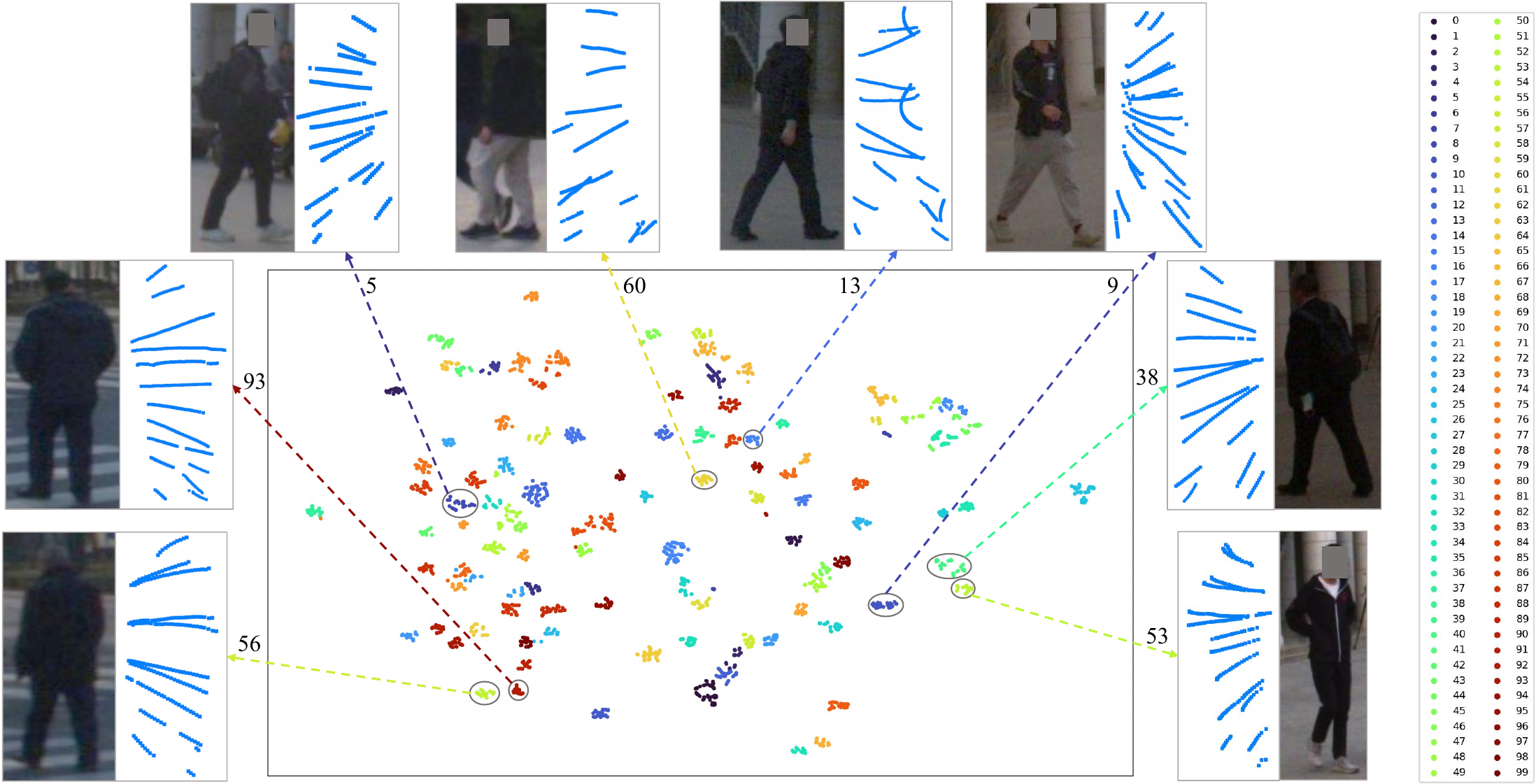}
	\end{center}
	\caption{The feature distributions of LReID test set extracted by ReID3D without pre-training (B-ReID3D). The images and corresponding point clouds are from the same pedestrians in Fig.~\ref{fig:tsne_tcl}. The features of pedestrians with similar appearances and those under low light are dispersed in the feature space, making them distinctive.}
	\label{fig:tsne_reid3d}
\end{figure*}

\begin{figure*}
	\begin{center}
		\includegraphics[width=0.9\linewidth]{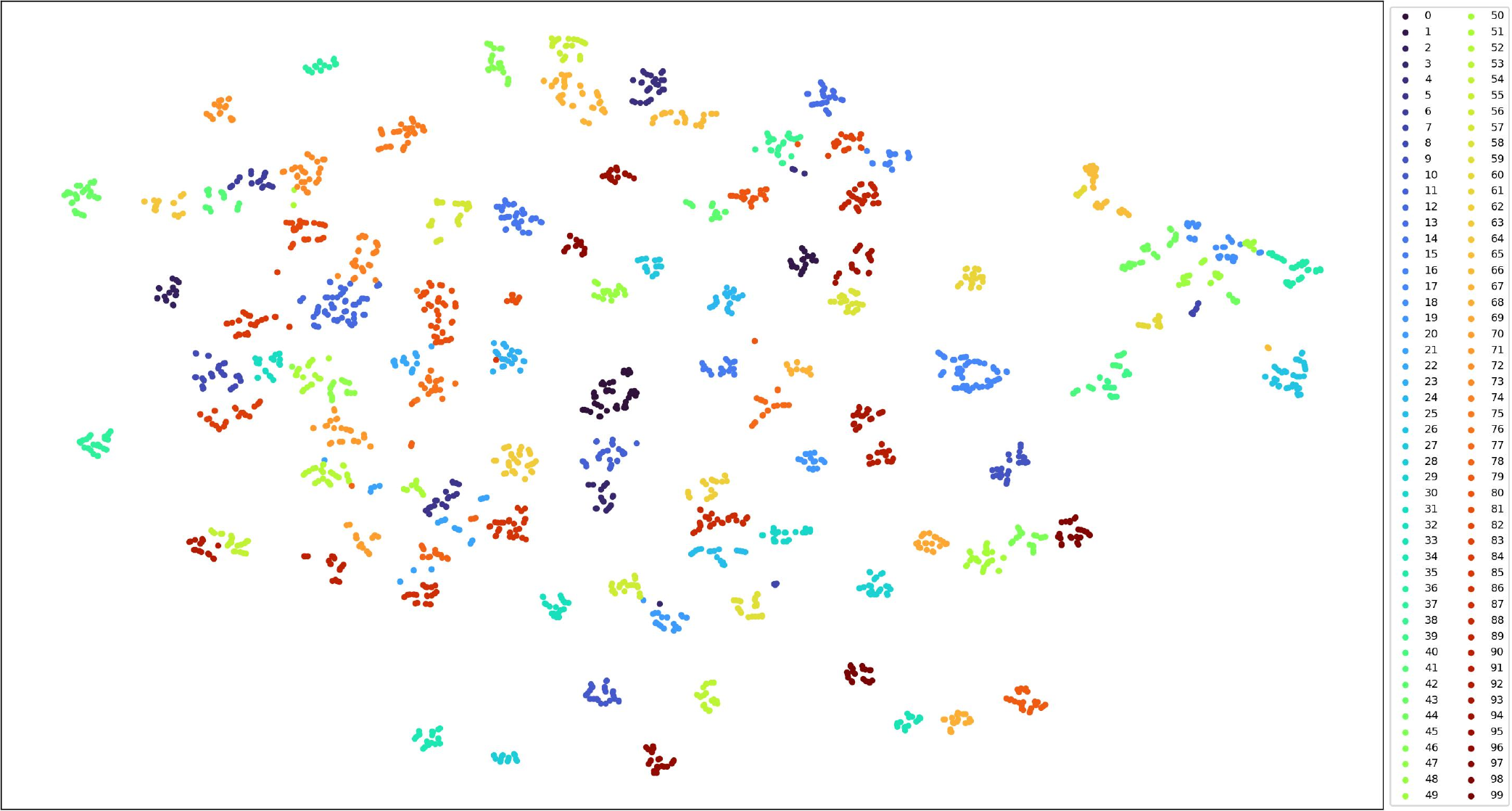}
	\end{center}
	\caption{The feature distribution of LReID test set extracted by Point Transformer~\cite{zhao2021point} with the same temporal fusion module.}
	\label{fig:tsne_pt}
\end{figure*}

To compare different point cloud encoders, we replace GCEE in B-ReID3D with Point Transformer~\cite{zhao2021point} and visualize its feature distribution, as shown in Fig.~\ref{fig:tsne_pt}. Compared to Fig.~\ref{fig:tsne_reid3d}, the features extracted by Point Transformer from samples of the same pedestrians are not effectively aggregated, and there are more scattered feature points distributed in the feature space. Additionally, the feature distances from different pedestrians lack sufficient distinctiveness. Therefore, Point Transformer exhibits a lesser capability in extracting features of pedestrian point clouds compared to GCEE.

